\documentclass[10pt,twocolumn,letterpaper]{article}

\usepackage[pagenumbers]{wacv} 
\usepackage{graphicx}
\usepackage{amsmath}
\usepackage{amssymb}
\usepackage{booktabs}
\usepackage{booktabs, multirow} 
\usepackage{soul}
\usepackage[table]{xcolor} 
\usepackage{changepage,threeparttable} 

\usepackage[pagebackref,breaklinks,colorlinks]{hyperref}
\usepackage[capitalize]{cleveref}
\crefname{section}{Sec.}{Secs.}
\Crefname{section}{Section}{Sections}
\Crefname{table}{Table}{Tables}
\crefname{table}{Tab.}{Tabs.}

\usepackage{graphicx}
\usepackage{amsmath}
\usepackage{amssymb}
\usepackage{booktabs}
\usepackage{rotating}
\usepackage{multirow}
\usepackage{tablefootnote}
\definecolor{LightCyan}{rgb}{0.8,1,1}
\definecolor{LightGreen}{rgb}{0.6,0.9,0.7}
\usepackage{xcolor, soul}
\sethlcolor{LightGreen}
\colorlet{LightCyan}{LightCyan}
\usepackage{colortbl}

\begin{document}

\title{\textsc{StyLIP}: Multi-Scale Style-Conditioned Prompt Learning\\for CLIP-based Domain Generalization}

\author{Shirsha Bose$^{1}\thanks{equal contribution}$ \and \hspace{-0.4cm}Ankit Jha$^{2*}$ \and \hspace{-0.4cm}Enrico Fini$^{3}$ \and \hspace{-0.3cm}Mainak Singha$^{2*}$ \and \hspace{-0.3cm}Elisa Ricci$^{3}$ \and \hspace{-0.4cm}Biplab Banerjee$^{2}$
\and
$^{1}$Technical University of Munich, Germany\and
$^{2}$Indian Institute of Technology Bombay, India\and
$^{3}$University of Trento, Italy
\and
{\tt\small shirshabosecs@gmail.com, ankitjha16@gmail.com, 
	enrico.fini@unitn.it}\and
 {\tt\small 
	mainaksingha.iitb@gmail.com, 
	e.ricci@unitn.it, getbiplab@gmail.com}}

\maketitle
\thispagestyle{empty}

\begin{abstract}
Large-scale foundation models, such as CLIP, have demonstrated impressive zero-shot generalization performance on downstream tasks, leveraging well-designed language prompts. However, these prompt learning techniques often struggle with domain shift, limiting their generalization capabilities.
In our study, we tackle this issue by proposing \textsc{StyLIP}, a novel approach for Domain Generalization (DG) that enhances CLIP's classification performance across domains. Our method focuses on a domain-agnostic prompt learning strategy, aiming to disentangle the visual style and content information embedded in CLIP's pre-trained vision encoder, enabling effortless adaptation to novel domains during inference.
To achieve this, we introduce a set of style projectors that directly learn the domain-specific prompt tokens from the extracted multi-scale style features. These generated prompt embeddings are subsequently combined with the multi-scale visual content features learned by a content projector. The projectors are trained in a contrastive manner, utilizing CLIP's fixed vision and text backbones.
Through extensive experiments conducted in five different DG settings on multiple benchmark datasets, we consistently demonstrate that \textsc{StyLIP} outperforms the current state-of-the-art (SOTA) methods.
\end{abstract}

\section{Introduction}
Advancements in large-scale vision and language models, such as CLIP \cite{clip} and ALIGN \cite{align}, have made remarkable progress in computer vision tasks. These models employ contrastively trained vision and text encoders to capture semantically meaningful concepts in a shared embedding space. They demonstrate impressive zero-shot generalization performance using text prompts like \texttt{A photo of a [CLS]}. However, designing an optimal prompt is challenging, and recent studies focus on data-driven prompt optimization \cite{coop}. Despite their success, prompt learning is limited to the training data distribution and is susceptible to domain shift \cite{csurka2017domain}. Domain shift, common in real-world applications, poses challenges as deep learning models are sensitive to differences between training and test data distributions \cite{batchnorm}. To tackle this, researchers explore Domain Generalization (DG) \cite{dg-survey, l2g, mldg}, which aims to learn a domain-agnostic representation from multiple datasets sourced from different domains for application to novel target domains. Traditional DG techniques rely on vision encoders trained exclusively on image data \cite{l2a-ot, ddaig, mixstyle}. Recent efforts combine foundation models with prompt engineering \cite{clip, coop} to bridge the semantic gap, but their practical applicability in DG settings requires further exploration.

\begin{figure}
    \centering
    \includegraphics[width=0.49\columnwidth]{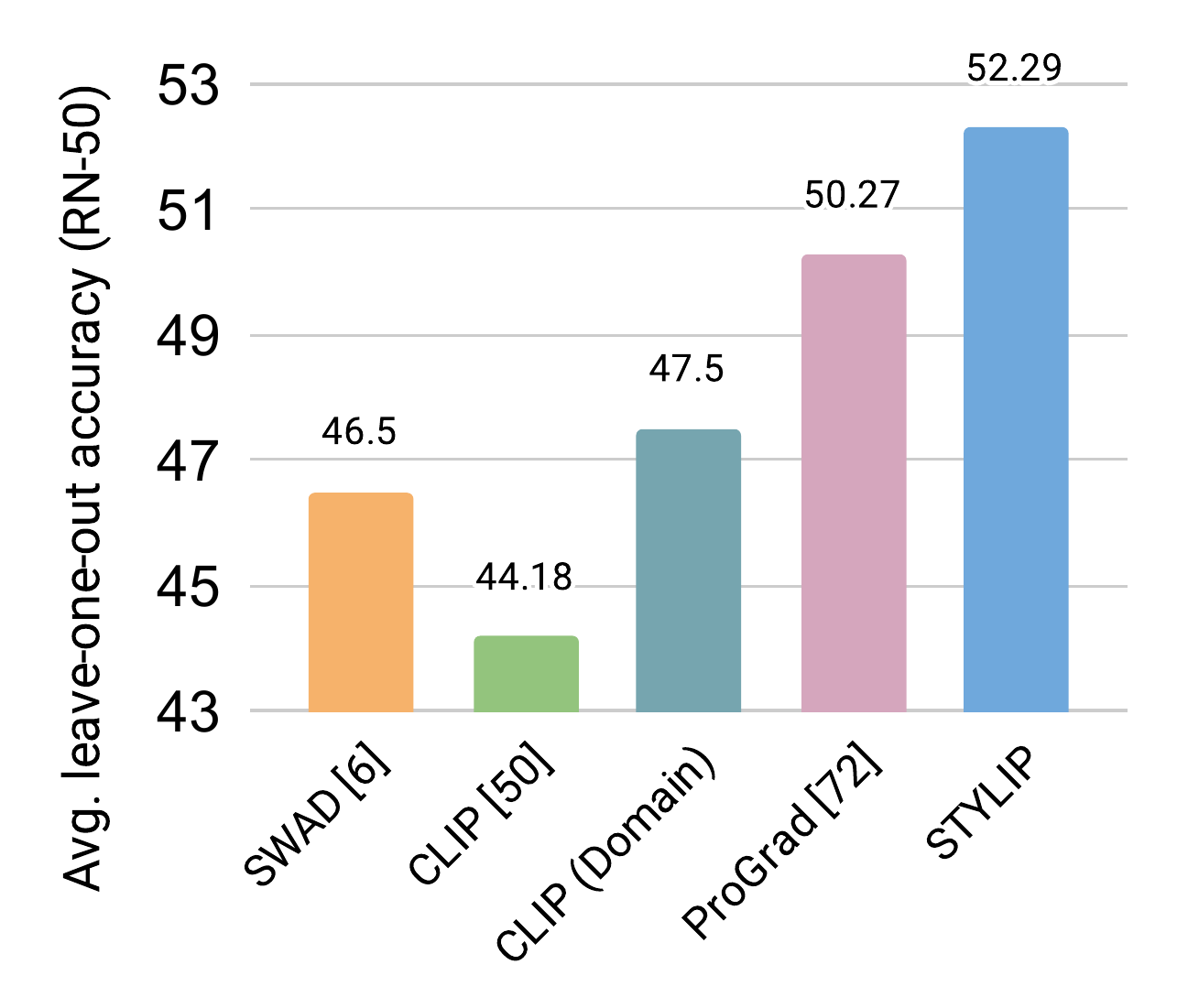}
    \includegraphics[width=0.49\columnwidth]{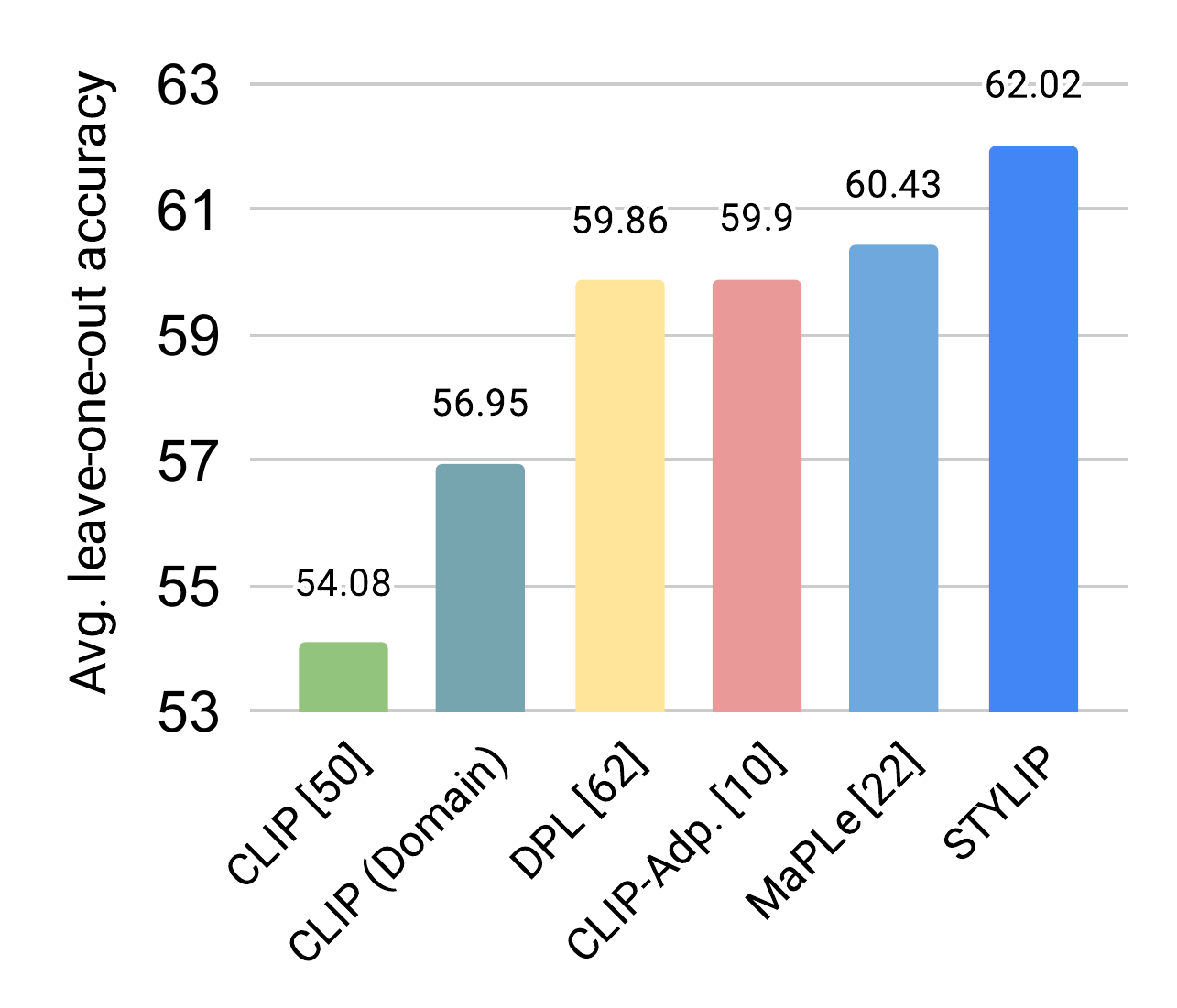}
    \vspace{-0.4cm}
    \caption{The performance of different DG techniques for the DomainNet \cite{domain_net} dataset for both the RN-50 and ViT backbones. Zero-shot CLIP falls behind SOTA traditional DG models like SWAD \cite{swad}, indicating poor generalization. However, incorporating domain identifiers in prompts boosts baseline CLIP's performance. Finally, \textsc{StyLIP} outperforms the previous best prompting techniques \cite{prograd, clip-adapter, maple} substantially, highlighting the importance of style and content disentanglement in the prompts for DG tasks.}
    \vspace{-0.5cm}
    \label{fig:teaser1}
\end{figure}

In this paper, we focus on a more challenging setting where significant visual variations exist across different domains, unlike existing prompting methods \cite{coop, cocoop, prod, prograd} that evaluate CLIP's generalization capabilities on datasets with limited domain shift (e.g., variants of ImageNet \cite{krizhevsky2012imagenet}). Fig. \ref{fig:teaser1} illustrates the average multi-source DG performance on DomainNet \cite{domain_net}, where zero-shot CLIP \cite{clip} underperforms compared to the best traditional DG model, SWAD \cite{swad}, by approximately 2.5\%. Using a domain-conditional prompt (\texttt{A [Domain] of a [CLS]}) boosts CLIP's accuracy by nearly 3\%, highlighting the importance of a representative prompt for DG. \textit{However, domain-level annotations are not always available, and the static domain name may not capture the style properties characterizing the domains \cite{mancini2018boosting}.}
Existing specialized prompt-tuning techniques \cite{amortized, prograd, cocoop} improve CLIP's performance (Fig. \ref{fig:teaser1}), but their effectiveness for DG is uncertain as prompts refined from random vectors may not effectively encode domain knowledge \cite{cocoop, prograd}. Zhang et al. \cite{amortized} propose a domain-prompt initialization strategy based on batch statistics of visual features but overlook important lower-level style characteristics and consider domain-level supervision. Another recent approach \cite{niu2022domain} learns prompts from CLIP without utilizing visual samples from the source domains but incorporates textual domain knowledge.

These discussions highlight a research gap in learning prompts that account for unknown domain shifts without explicit domain identifiers. We argue that leveraging visual features in such scenarios is crucial, along with dynamically incorporating object-level variations into the prompts to aid in cross-domain generalization tasks \cite{cocoop}. Motivated by these considerations, our research question is \textit{ whether we can utilize CLIP's vision backbone to encode image style and content information for learning domain and instance-aware prompts to address DG.}

\textbf{Our proposed \textsc{StyLIP}:} We introduce \textsc{StyLIP}, a novel generic prompt tuning strategy for CLIP that addresses these challenges. Our approach aims to enhance the prompts' understanding of class concepts by conditioning them on domain and content information derived from CLIP's visual space. To achieve this, we leverage the hypothesis that instance-wise feature statistics from intermediate layers of an image encoder capture the visual domain information \cite{instyle}. We extract mean and standard deviation values from CLIP's intermediate feature map outputs and utilize a set of \textsc{style projectors} to learn domain-specific tokens in the prompts. Unlike existing models such as \cite{coop, cocoop, prograd} that learn prompt token embeddings from ad-hoc sentences, our approach benefits from using style features at different scales, which leads to improved domain-aware prompts and better prompt initialization. 

Additionally, we propose to incorporate image content information into the prompt embeddings to capture object-level variations and avoid overfitting to the training classes, which is particularly important in DG scenarios concerning disjoint training and test classes. While Zhou et al. \cite{cocoop} addresses this issue by adding high-level semantic features from CLIP's vision encoder to prompt tokens, we consider a DG setting where the distributions of training and test classes differ \cite{cumix}. To achieve this, we combine visual feature responses from different layers of CLIP's vision encoder and aggregate them through a \textsc{content projector}. By encoding mid- to high-level image characteristics that are more generic across categories \cite{zheng2016good}, we aim to enhance transferability across domains/classes. Unlike the literature \cite{maple, cocoop}, we propose a learnable fusion network to aggregate these visual features with the final prompt embeddings obtained previously.

\noindent\textbf{Contributions:} We highlight our major contributions as: \\
- We introduce \textsc{StyLIP}, a domain-unified prompt learning strategy that leverages CLIP's frozen vision encoder to extract the domain and content information from an image and deploy them in prompt learning through light-weight learnable style and content projectors.\\ 
- We acquire prompt tokens from visual style features at various scales, facilitating the consolidation of hierarchical domain knowledge, thereby assisting in generalization across different domains. Moreover, we incorporate multi-scale visual content information into prompt embeddings, effectively mitigating overfitting and promoting generalization across different categories.\\
- We showcase the performance of \textsc{StyLIP} for multiple datasets on five major DG tasks: i) single-source and multi-source DG, ii) cross-dataset DG, iii) in-domain base to novel class DG, and iv) cross-domain base to novel class DG, a novel task we introduce in the context of prompting. Experimentally, \textsc{StyLIP} outperforms the competitors in all tasks at least by $0.2-4 \%$.
\textit{To our knowledge, ours is the first attempt to extensively study the DG problem using CLIP.}

\section{Related Works}
\label{sec:related_works}
\noindent \textbf{Domain generalization}. The DG problem has different variations. Single-source DG \cite{sdg1, sdg2} trains with one domain, while multi-source DG \cite{l2a-ot, style-neophile, lostdg} considers training multiple domains simultaneously. In a closed-world setting, where the label set is shared across domains, DG approaches commonly address domain shift. Heterogeneous DG \cite{hetdg1, hetdg2} faces additional challenges due to different labels between the source and target domains.
Previous research on DG proposed methods such as domain alignment losses \cite{li2020domain, wang2021respecting, jia2020single, MMD}, self-supervised learning \cite{jigen}, ensemble learning \cite{xu2014exploiting}, domain-specific networks \cite{mancini2018best}, and meta-learning \cite{learningtolearn}. However, these methods often require more training domains, which can influence DG performance. To overcome this, novel pseudo-domains have been generated using domain augmentation approaches \cite{sfa, l2a-ot, style-neophile, mixstyle}. In single-source DG models \cite{sdg1, sdg2, sdg3}, diverse styles can be synthesized by perturbing the source domain through entropy maximization, meta-learning, and adversarial learning. Conversely, methods for heterogeneous DG \cite{hetdg3, hetdg4, l2a-ot} aim to improve model generalizability for novel tasks.

DPL \cite{amortized} used CLIP \cite{clip} for multi-source DG by inferring domain information from batch-wise visual features. However, DPL doesn't fully leverage CLIP's ability to extract domain-specific artifacts and can overfit with small batches due to challenges in obtaining an unbiased style estimate. Researchers have explored domain invariant prompts \cite{niu2022domain, li2022learning} through text-based source domain knowledge or image patches for prompt input in ViT models, similar to VPT \cite{vpt}. \textit{Our \textsc{StyLIP} approach differs from \cite{niu2022domain, amortized, li2022learning} by considering style features at different visual encoder levels to learn individual prompt tokens and exploring multi-scale visual features in prompt learning, which have been successful in various DG tasks.}

In a recent study, Cumix \cite{cumix} combines DG with the notion of zero-shot learning \cite{zs_good_bad} for the recognition of new domains and classes. The following research investigated the use of structured multimodal information \cite{chandhok2021structured} or disentangled feature learning \cite{mangla2021context} for similar aims. \textit{Our proposed experimental setup for a base to novel class generalization is identical; however, we are interested in analyzing the performance of the prompting techniques for VLMs in this respect, contrary to the more ad-hoc models mentioned above}.

\noindent \textbf{Prompt tuning for vision-language models (VLMs)}.
VLMs have gained attention in language processing and computer vision \cite{lee2018pre, radford2019language, bommasani2021opportunities, bossard2014food, singha2023ad,helber2019eurosat,singha2023applenet}. These models utilize task-centric textual descriptions for visual data \cite{henaff2020data, huynh2020fine}. Earlier prompting strategies were manual but later works focused on prompt learning.
CoOp \cite{coop} optimized unified and class-specific prompts through back-propagation. CoCoOp \cite{cocoop} addressed CoOp's generalization issue through input-conditioned prompt learning. CLIP-adapter \cite{clip-adapter} proposed fine-tuning feature adapters in visual and language branches.
ProGrad \cite{prograd} prevents knowledge forgetting from the foundation model. TPT \cite{tpt} utilizes consistency among multiple image views for supervision. Probabilistic and variational models \cite{prod, varprompt} learn prompt distributions to match visual feature spreads. LASP \cite{LASP} improves the learned prompt via text-to-text cross-entropy loss. MaPle \cite{maple} enhances compatibility between CLIP encoders at different levels. However, these approaches are not tailored to deal with multi-domain data.
\textit{In opposition, we introduce the notion of visual content-style disentanglement for prompt learning for DG tasks using CLIP.}

\begin{figure*}
    \centering
    \includegraphics[width=\linewidth]{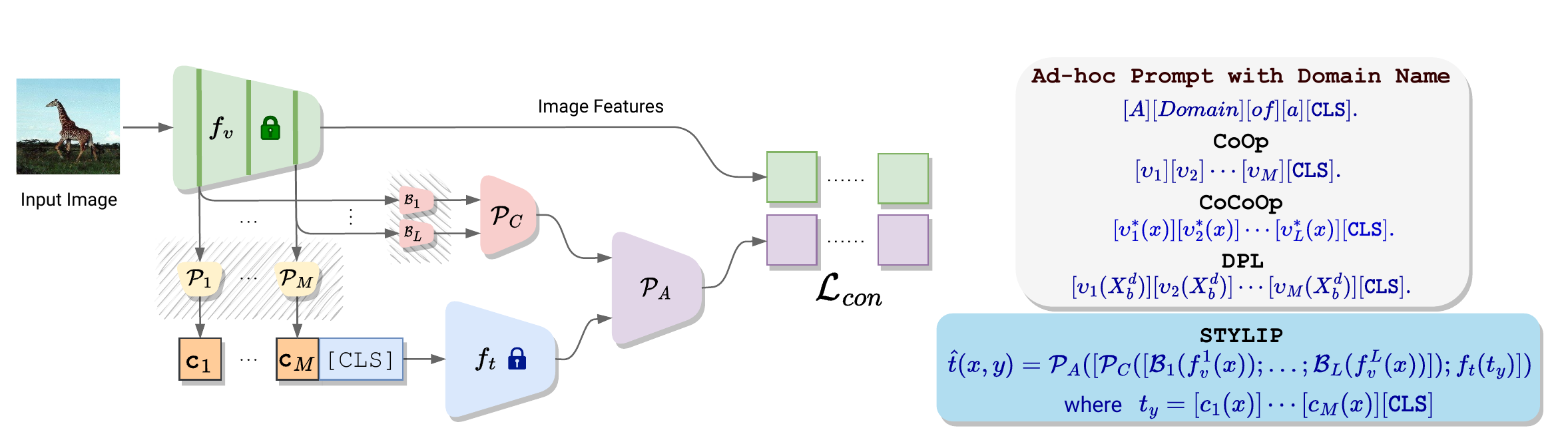}
    \vspace{-0.8cm}
    \caption{\textsc{StyLIP} generates a joint embedding space for image and prompt embeddings, leveraging style and content information extracted from the vision encoder $f_v$. The projectors $\{\mathcal{P}_m\}_{m=1}^M, \{\mathcal{B}_l\}_{l=1}^L$, and $\mathcal{P}_C$ adapt these features for the target task. For each image $x$, the style features from layer $m$ of $f_v$ are used to learn the $m^{th}$ token of the prompt, $c_m(x)$, through the style projector $P_m$. Similarly, the multi-scale content features are aggregated using $\mathcal{P}_C$ after dimensionality reduction by the bottleneck networks ${\mathcal{B}_l}$. The fusion projector $\mathcal{P}_A$ computes the classification weights. Unlike existing prompt tuning techniques (CoOp, CoCoOp, and DPL), \textsc{StyLIP} differs in its approach. $v_m$ represents the prompt learner from a random vector, and $X_b^d$ denotes a batch of samples for domain $d$. }
    \vspace{-0.4cm}
    \label{fig:architecture}
\end{figure*}

\section{Proposed Methodology}
\label{sec:Proposed_Methodology}
\subsection{Problem and notation}
The DG problem involves $\mathcal{N}$ labelled source domains $\mathcal{S}^i = \{x_i^k, y_i^k\}_{k=1}^{n_i} \approx P^{\mathcal{S}^i}_{data}$, $1 \leq i \leq \mathcal{N}$, where $x_i \in \mathcal{X}^i$, $y_i \in \mathcal{Y}$, and $P^{\mathcal{S}^i}_{data}$ denote the input data, label, and the joint distribution concerning the data and the label space, respectively. Furthermore, $P^{\mathcal{S}^i}_{data} \neq P^{\mathcal{S}^j}_{data}$ $\forall i,j \in \{1, 2, \cdots, \mathcal{N}\}$, indicating that the source domains are mutually distinct. We call the setting single-source DG if $\mathcal{N}=1$, else it is known as multi-source DG.
The goal is to train a model $f:\mathcal{X} \rightarrow \mathcal{Y}$ given $\mathcal{S} = \{\mathcal{S}^i\}_{i=1}^{\mathcal{N}}$, which is expected to generalize for a novel target domain $\mathcal{S}^{\mathcal{N}+1} = \{x_t^k, y_t^k\}_{k=1}^{n_t}$ unseen during training with $x_t \in \mathcal{X}^t$ and $y_t \in \mathcal{Y}^t$ and $P^t_{data}$ denotes the target distribution which is different from the source distributions. Typically, we consider a closed-set setting where $\mathcal{Y} \cup \mathcal{Y}^t = \mathcal{Y} \cap \mathcal{Y}^t$. Also, for the base to new class generalization setting, we consider $\mathcal{Y} \cap \mathcal{Y}^t = \emptyset$.

\subsection{The \textsc{StyLIP} model}
\label{sec:details}
Here, we introduce \textsc{StyLIP}, a novel approach for DG based on CLIP \cite{clip}.
\textsc{StyLIP} leverages CLIP's frozen vision encoder ($f_v$) and text encoder ($f_t$), trained on a large volume of image-text pairs (see Fig. \ref{fig:architecture}). $f_v$ that transforms an input image into a feature embedding vector can be implemented with different architectures: in our experiments (see Section \ref{sec:results}), we consider ResNet-50 (RN50) \cite{resnet}, and ViT-B/16 \cite{vit}. $f_t$ is built upon a Transformer \cite{transformer}: it is provided with an input of a sequence of word tokens and converts them into a vectorized representation. 

As stated, \textsc{StyLIP} seeks to utilize the multi-scale visual features extracted from different levels of $f_v$ to estimate the style and content primitives and further channel them in learning a generic prompt space regarding a concept. Typically, high-level representations of the deepest layer of a vision encoder tend to capture the abstract object semantics suitable for classification but suffer from a lack of description of local patterns like oriented edges or local shapes \cite{zheng2016good}. Therefore, the set of characteristics obtained from multiple levels is deemed more transferable between tasks than the high-level features alone. Similarly, the instance-wise feature statistics calculated from multiple layers of the encoder capture different levels of style, e.g., the texture in the top layers usually has larger granularity than those in the bottom layers \cite{babenko2014neural}.

To model a continuous prompt embedding space using these multilevel visual features, \textsc{StyLIP} (see Fig. \ref{fig:architecture}) adopts a set of projector networks on top of $f_v$ and $f_t$: a set of $M$ style projectors $\{\mathcal{P}_m\}_{m=1}^M$ to encode domain characteristics into $M$ prefix tokens $\{c_m\}_{m=1}^M$, a content projector $\mathcal{P}_C$ to encode feature responses from all the $L$ encoder layers of $f_v$ after reducing their dimensions using bottleneck layers $\{\mathcal{B}_l\}_{l=1}^L$,
and a fusion projector $\mathcal{P}_A$.
We discuss the structure of the proposed projectors in detail below.


\noindent\textbf{Embedding multi-level style information into prompt tokens:} For calculating the style features, let us consider the vector $\mathcal{F}_l(x) = [\vec{\mu_l}(x); \vec{\sigma_l}(x)]$ denoting the channel-wise mean and standard deviation of the feature map outputs from the $l^{th}$ layer ( $1 \leq l \leq L$) of $f_v$, also indicated as $f_v^l(x)$. Here, 
$[-;-]$ denotes the concatenation operation. Specifically, if $f_v^l(x)$ is of dimensions $W \times H \times C$ (height, width, and depth dimensions), the statistics corresponding to the $c^{th}$ feature map $f_v^{l_c}, $ ($\mu_l^c, \sigma_l^c$), are calculated as:
\vspace{-3mm}
\begin{equation}
\label{mu_sigma1}
\begin{aligned}
    \centering
    \mu_l^c &= \frac{1}{WH} \underset{w,h=1}{\overset{W,H}{\sum}} f_v^{l_c}(x)_{w,h}
\end{aligned}
\end{equation}
\vspace{-1mm}
\begin{equation}
\label{mu_sigma2}
\begin{aligned}
    \centering
    \sigma_l^c &= \sqrt{\underset{w,h=1}{\overset{W,H}{\sum}} (f_v^{l_c}(x)_{w,h} - \mu_l^c)^2}
\end{aligned}
\end{equation}

In the simplest case, when the context length $M$ equals the number of encoder layers $L$, we seek to learn the $m^{th}$ context vector $c_m$ from $\mathcal{F}_m(x)$. 
Considering that the dimensions of $\mathcal{F}_m(x)$s are inconsistent and
to appropriately input $\mathcal{F}_m$ into the text encoder $f_t$, we deploy the style projectors $\{\mathcal{P}_m\}_{m=1}^M$ and compute $c_m(x) = \mathcal{P}_m(\mathcal{F}_m(x))$, \textit{i.e.} the $m^{th}$ context vector for the text prompt. We define 
\vspace{-1.25mm}
\begin{equation} \centering
t_y = [c_1(x)] [c_2(x)] \cdots [c_M(x)] [CLS_y]
\label{eq:prompt}
\vspace{-.5mm}
\end{equation}
as the prompt for $(x, y)$ where $[CLS_y]$ is the word embedding of label $y$. Finally, $f_t$ generates the embedding $f_t(t_y)$.

However, the context length in prompting is a hyper-parameter, meaning an $M$ different from $L$ may be preferred for a given task. To incorporate this flexibility in our prompt learning, we consider aggregation or replication of representations from $\{\mathcal{F}_l(x)\}_{l=1}^L$ depending on whether $M < L$ or $M > L$, respectively.

 
 \noindent\textbf{Supplementing the prompt embeddings with multi-scale image content features:} The paradigm of $t_y$ considers the information of the visual style of the images, but a static class embedding $CLS_y$ for all images with the label $y$ may limit its versatility. To further generalize the prompt embeddings, we propose to supplement $t_y$ with content image information. As discussed, we extract multiple complementary visual characteristics associated with an image by aggregating the multilevel feature responses obtained from the $L$ blocks of $f_v$.

One naive way to combine this multilevel information is by flattening the feature maps of individual blocks, followed by concatenation. However, this leads to a very high-dimensional vector representation compared to the dimensionality of $t_y$, undermining the effects of $t_y$ in the final classification weights. As a result, the contrastive task may lead to triviality. We propose reducing the feature maps' dimensions before concatenation as a remedy. This also shrinks the size of the inputs to $\mathcal{P}_C$, thus controlling its number of learnable parameters and the amount of information exchanged by the two encoders.

 Precisely, given the $l^{th}$-layer feature maps $f^l_v(x)$, we perform $1 \times 1$ convolution followed by flattening using $\mathcal{B}_l$ to reduce the channel depth of $f^l_v(x)$ from original $C$ to $\hat{C} << C$, resulting in $\mathcal{B}_l(f_v^l(x)) \in \mathbb{R}^{WH\hat{C} \times 1}$. 
 Finally, we concatenate the $\mathcal{B}(f_v^l(x))$s to obtain $\hat{f}_v(x)$: 
 \vspace{-2.mm}
 \begin{equation}
 \centering
 \hat{f}_v(x) = [\mathcal{B}_1(f_v^1(x));\mathcal{B}_2(f_v^2(x));\cdots; \mathcal{B}_L(f_v^L(x))]
   \vspace{-2.mm}
 \end{equation}

 The content projector $\mathcal{P}_C$ learns the combined image embedding $\mathcal{P}_C(\hat{f}_v(x))$ through a linear transformation. To generate the classification weights for a given $(x, y)$, we first concatenate $\mathcal{P}_C(\hat{f}_v(x))$ with $f_t(t_y)$ and transform the aggregated information through the fusion projector $\mathcal{P}_A$ to obtain $\hat{t}(x, y)$ as follows:
  \vspace{-2.mm}
\begin{equation}
    \centering
    \hat{t}(x, y) = P_A([\mathcal{P}_C(\hat{f}_v(x));f_t(t_y)])
      \vspace{-2.mm}
\end{equation}

\subsection{Training and inference}
 The projectors are trained using a contrastive loss $\mathcal{L}_{con}$ between $\hat{t}(x, y)$ and the image features obtained from the final embedding layer of $f_v$, i.e., $f_v(x)$, as follows:
\vspace{-0.25cm}
 \begin{equation}
      \centering
     \mathcal{L}_{con} = \underset{\substack{\{\mathcal{P}_m\}_{m=1}^M, \\\{\mathcal{B}_{l}\}_{l=1}^L, \mathcal{P}_C, \mathcal{P}_A}}{\arg \min} \underset{{(x, y) \sim P^{\mathcal{S}}_{data}}}{\mathbb{E}}\ - \log (p(\hat{t}(x, y)\|x))
     \vspace{-0.15cm}
 \end{equation}
\noindent where $P^{\mathcal{S}}_{data}$ is the joint data distribution of $\mathcal{S}$ and
\vspace{-0.15cm}
\begin{equation}
     \centering
    p(\hat{t}(x^k, y^k)\|x^k) = \frac{e^{\delta(\hat{t}(x^k, y^k), f_v(x^k))/\tau}}{\underset{n \in \mathcal{Y}}{\sum}e^{\delta(\hat{t}(x^k,n), f_v(x^k))/ \tau}}
    \vspace{-0.15cm}
\end{equation}
$\delta$ defines the cosine similarity and $\tau$ is the temperature hyperparameter. 
The contrastive loss synergistically maximizes the similarity between the image and the correct class prompt embeddings while minimizing the similarity between the image and all the opposing classes.

During inference, we calculate the compatibility between $f_v(x_t)$ and the prompt embeddings for all classes in $\mathcal{Y}^t$. The class with the highest compatibility is selected as:
\vspace{-0.15cm}
\begin{equation}
    \hat{y}_t = \underset{n \in \mathcal{Y}^t}{\arg \max} \: p(\hat{t}(x_t,n)\|x_t)
\end{equation}

\section{Experimental Results}
\label{sec:results}

\noindent\textbf{Datasets:} We evaluate \textsc{StyLIP} over five benchmark datasets for multi-source and single-source domain generalization, namely \texttt{Office-Home} \cite{officehome}, \texttt{PACS} \cite{dbadg}, \texttt{VLCS} \cite{vlcs}, \texttt{Digits-DG} \cite{l2a-ot} ], and \texttt{DomainNet} \cite{domain_net}. We further analyze the performance of \textsc{StyLIP} for cross-dataset generalization, where \textsc{StyLIP} is trained on \texttt{ImageNet} \cite{krizhevsky2012imagenet} and tested on ten other different datasets \cite{coop}. Detailed descriptions of the datasets are provided in \textsc{supple.}

\begin{figure}[t!]
    \centering
    \includegraphics[width=7.6cm]{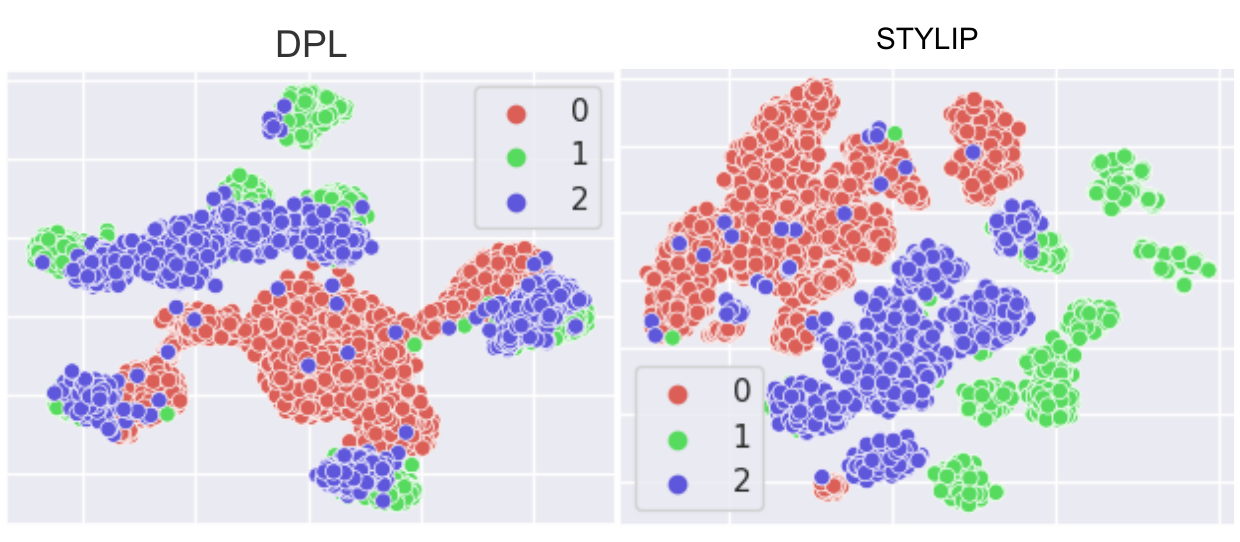}
    \caption{t-SNE visualization of the prompt embedding outputs from $f_t$ for DPL \cite{amortized} and \textsc{StyLIP} for multi-DG on PACS dataset. All the domains are highly clustered in \textsc{StyLIP}.}
    \label{fig:tsne}
\end{figure}

\noindent\textbf{Implementation, training, and evaluation protocols:} We implement the projectors in $(\mathcal{P}_C, \mathcal{P}_A, \{\mathcal{P}_m\}_{m=1}^M)$ as single dense layers. We train the model with Adam optimizer \cite{kingma2014adam} with a learning rate of $2e-2$ and betas $(0.9, 0.999)$. 
 We consider a context length of four for all the experiments following \cite{coop, cocoop}. For RN50, we consider the feature map outputs from the four convolution stages to extract the style and content features, hence $L=M=4$. For ViT-B/16, we obtain the embedding outputs from the $L=12$ encoder layers. We further average the features for every three consecutive layers of $f_v$ in a bottom-up manner without overlap to generate four intermediate feature representations, which are subsequently used to produce four distinct domain information vectors to be passed to $\{\mathcal{P}_m\}_{m=1}^4$. As we show in Fig. \ref{fig:tsne}, the feature statistics capture the domain information in ViT-based prompt learning for \textsc{StyLIP}, similar to RN50.
 We fix the number of output channels of the bottleneck $\hat{C}=3$ using cross-validation, where $10 \%$ images from each source domain are treated as the validation set. We further ablate $\hat{C}$ in Section \ref{sec:ablation} to check other architecture choices.
Finally, we consider a mini-batch size of $4$ for DomainNet and Office-Home, while it is $8$ for the other datasets, and we train the model for $10$ epochs. We report the average top-1 classification performance on $\mathcal{S}^{\mathcal{N}+1}$ over three different executions. \textit{In terms of model complexity, \textsc{StyLIP} is extremely light-weight and consists of $0.18 \%$ more parameters than CoOp and CoCoOp and $0.06 \%$ more parameters than MaPLe.}

\noindent\textbf{Baselines:} We consider three types of methods for comparison to check the generalizability of pre-trained CLIP features and that of the prompting strategies.
Our baseline is \textit{zero-shot CLIP} with the prompt as `\texttt{A Photo of a [CLS]}'. We also include \textit{domain name} in the prompt as `\texttt{A [Domain] of a [\textit{CLS}]}'. We use CLIP features to train a linear classifier, which we term \textit{Linear Probing}. Furthermore, we deploy these features in conjunction with the benchmark DG technique of \textit{CROSSGRAD} \cite{cross-grad}, where we put the learnable networks on top of frozen CLIP for back-propagation training. From the traditional DG literature, we report the performance of \textit{SWAD} \cite{swad} for Multi-DG and SagNet \cite{nam2021reducing} and DSBF \cite{dsbf} for Single-DG, respectively. Furthermore, we choose to compare \textsc{StyLIP} with existing prompt learning techniques including \textit{CoOp} \cite{coop}, \textit{CoCoOp} \cite{cocoop}, \textit{CLIP-Adapter} \cite{clip-adapter}, \textit{DPL} \cite{amortized}, \textit{ProGrad} \cite{prograd}, \textit{VPT} \cite{vpt}, \textit{CSVPT} \cite{li2022learning}, \textit{MaPLe} \cite{maple} and, \textit{TPT} \cite{tpt}, etc.

Finally, we evaluate three variants of \textsc{StyLIP} intending to ablate the individual components of our approach: \textbf{i)}  $\{\mathcal{P}_m\}_{m=1}^M$ are trained from random vectors (similar to CoOp), but we consider the multi-scale content feature learning of \textsc{StyLIP} through $(\{\mathcal{B}_l\}_{l=1}^L, \mathcal{P}_A, \mathcal{P}_C)$, respectively. (\texttt{\textsc{StyLIP}-con}). This establishes the importance of including the visual style information in the prompt tokens. \textbf{ii)} the model without content features and $(\{\mathcal{B}_l\}_{l=1}^L, \mathcal{P}_C)$ but with $\{\mathcal{P}_m\}_{m=1}^M$ (\texttt{\textsc{StyLIP}-sty}). This is to verify the importance of the multi-scale content features, and \textbf{iii)} the version of \textsc{StyLIP} where the features of the deepest layer of $f_v$ are used for the content branch together with $\mathcal{B}_L$ only (\texttt{\textsc{StyLIP}*}). This is to assess the importance of the multi-scale content features over the single-scale high-level visual content properties as used in the literature \cite{cocoop}.

\begin{table}[!ht]
    \scriptsize{
    \begin{center}
    
    \caption{Comparison of our proposed \textsc{StyLIP} with the state-of-the-art methods on PACS, VLCS, Digits-DG, Office-Home, and DomainNet datasets for multi-source DG in terms of mean leave-one-out performance. $\dagger$ uses a different backbone than CLIP. (In $\%$) $^\#$\tablefootnote{Methods are trained for a large number of epochs such as VPT \cite {vpt} trains for $100$ epochs, whereas \textsc{StyLIP} is trained only with $10$ epochs.}\hl{Methods perform fine-tuning the visual backbones.}}
    \vspace{-0.32cm}
    \scalebox{0.77}{
    \begin{tabular}{llccccc}
    \toprule
    \multicolumn{1}{c}{\textbf{Backbone}} &\multicolumn{1}{l}{\textbf{Method}} & \multicolumn{1}{c}{\textbf{PACS}}& \multicolumn{1}{c}{\textbf{VLCS}}& \multicolumn{1}{c}{\textbf{Off.Home}}&\multicolumn{1}{c}{\textbf{Dig.DG}}&\multicolumn{1}{c}{\textbf{Dom.N.}}\\ 
    \midrule
         \multirow{17}{*}{\rotatebox{90}{CLIP RN50}} & SWAD$^{\dagger}$ \cite{swad} &88.10 &  79.10& 70.60&-&46.50  \\
       
        \cmidrule(lr){2-7}
         &  Lin. Probing & 91.65& 79.48 & 70.17 &62.22&46.10\\
        ~ & CROSSGRAD \cite{cross-grad} & 91.56 & 79.63&    70.47&62.98&45.64\\ 
        &ZS-CLIP \cite{clip}& 90.32 & 76.43& 66.75&56.41&44.18\\ 
       & ZS-CLIP + DN&  91.86 & -&67.93&-&47.50\\

        ~ & CoOp \cite{coop}& 92.28 & 81.87& 71.65&73.11& 49.71\\ 
        ~ & CoCoOp \cite{cocoop} & 91.64& 82.30 & 71.93&74.58& 50.16\\ 
        ~ & CLIP-Adapt. \cite{clip-adapter}&  92.08 & 82.35 & 72.18&73.79&50.25 \\
        ~ & DPL \cite{amortized} & 91.96 & 82.12 & 72.54&74.33&50.38\\ 
        
        ~ & ProGrad \cite{prograd} & 92.01 & 82.23& 71.85&74.45&50.27\\ 
        ~ & TPT \cite{tpt} & 92.16 & 82.39 &72.07&74.68&50.30\\ 
        
         &\cellcolor{LightCyan}\textsc{StyLIP}-con&	\cellcolor{LightCyan}92.35&	\cellcolor{LightCyan}84.07&	\cellcolor{LightCyan}73.89&\cellcolor{LightCyan}75.90&\cellcolor{LightCyan}51.43\\
        ~ & \cellcolor{LightCyan}\textsc{StyLIP}-sty  & \cellcolor{LightCyan}92.96  &\cellcolor{LightCyan}84.39& \cellcolor{LightCyan}74.22&\cellcolor{LightCyan}76.21&\cellcolor{LightCyan}51.80\\
        
        &	\cellcolor{LightCyan}\textsc{StyLIP}*&  \cellcolor{LightCyan}92.47&	\cellcolor{LightCyan}83.60&	\cellcolor{LightCyan}73.56&	\cellcolor{LightCyan}75.81&\cellcolor{LightCyan}51.63\\
        
        ~ & \cellcolor{LightCyan}\textsc{StyLIP} & \cellcolor{LightCyan}\textbf{93.59} & \cellcolor{LightCyan}\textbf{84.83} & \cellcolor{LightCyan}\textbf{74.80} & \cellcolor{LightCyan}\textbf{76.49}& \cellcolor{LightCyan}\textbf{52.29} \\ \midrule
        \multirow{16}{*}{\rotatebox{90}{CLIP ViT-B/16}} & Lin. Probing  & 96.54  & 82.63& 80.43 &70.15&57.46\\ 
        ~ & CROSSGRAD \cite{cross-grad} &96.40  &83.76  & 80.55 &70.83&57.60\\ 
        & ZS-CLIP \cite{clip}&95.81 & 80.57& 78.57&65.79&54.08\\ 
        & ZS-CLIP + DN&96.30  &- &79.10 &-&56.95\\
        ~ & CoOp \cite{coop}& 97.00 & 82.98&	81.12 &76.41&59.52\\ 
        ~ & CoCoOp \cite{cocoop}& 96.73 & 83.59&	80.70&78.49&59.68\\ 
        ~ & CLIP-Adapt. \cite{clip-adapter}&96.41 & 84.32&82.23&77.86&59.90\\ 
        ~ & DPL \cite{amortized} & 97.07 & 83.99& 83.00 &77.32&59.86\\ 
        ~ & ProGrad \cite{prograd}&96.50 &83.82& 82.46 &78.26&59.65\\ 
        ~ & TPT \cite{tpt} & 96.99 &83.72&82.45&78.51&59.87 \\
        ~ & MIRO \cite{miro} &95.80 &83.60 &82.30 & - &57.20\\
        ~ & \hl{VPT}$^\#$\cite{vpt} &97.20 &84.90 &85.20 & - &59.80 \\
        ~ &\hl{CSVPT}$^\#$\cite{csvpt} &97.30 &84.90 &85.00 & - &60.00 \\
        ~ &\hl{DUPRG}$^\#$\cite{duprg} & 97.10& 83.90& 83.60& - &59.60\\
        ~ &\hl{MaPLe}$^\#$\cite{maple} & 97.56 & 85.12 & 83.35 & - & 60.43 \\
         &\cellcolor{LightCyan}\textsc{StyLIP}-con & \cellcolor{LightCyan}96.82&	\cellcolor{LightCyan}85.61& 	\cellcolor{LightCyan}83.90&\cellcolor{LightCyan}80.63&\cellcolor{LightCyan}61.51\\
         & \cellcolor{LightCyan}\textsc{StyLIP}-sty & \cellcolor{LightCyan}97.25  & \cellcolor{LightCyan}86.27&  \cellcolor{LightCyan}84.18 &\cellcolor{LightCyan}80.91&\cellcolor{LightCyan}61.77\\
         
         &	\cellcolor{LightCyan}\textsc{StyLIP}*&  \cellcolor{LightCyan}97.11&	\cellcolor{LightCyan}85.88&	\cellcolor{LightCyan}83.41&	\cellcolor{LightCyan}80.56&\cellcolor{LightCyan}61.39\\
         
         &\cellcolor{LightCyan}\textsc{StyLIP}&\cellcolor{LightCyan}\textbf{98.05}&\cellcolor{LightCyan}\textbf{86.94}  &\cellcolor{LightCyan}\textbf{84.63} &\cellcolor{LightCyan}\textbf{81.38}&\cellcolor{LightCyan}\textbf{62.02}\\ 
         \bottomrule
    \end{tabular}}
    \label{tab_PACS}
    \end{center}}
    \vspace{-0.6cm}
\end{table}

\begin{table}[!ht]
    \centering
    \scriptsize{
    \caption{Comparing \textsc{StyLIP} with CLIP-based SOTA methods for single-source DG on PACS, VLCS, and Office-Home datasets in terms of mean leave-all-but-one-domain-out performance. (In $\%$) }
    \vspace{-0.32cm}
    \scalebox{0.85}{
    \begin{tabular}{clccc}
    \toprule
   \multirow{2}[1]{*}{\textbf{Backbone}}&\multirow{2}[1]{*}{\textbf{Method}}&\multicolumn{1}{c}{\textbf{PACS}}&\multicolumn{1}{c}{\textbf{VLCS}}&\multicolumn{1}{c}{\textbf{Office Home}}\\
          \cmidrule(lr){3-3}\cmidrule(lr){4-4}\cmidrule(lr){5-5}
       \midrule
 &SagNet \cite{nam2021reducing} & 61.90 &-&68.00\\ 
 &DSBF \cite{dsbf} & 85.33 &-&63.91\\ 
 \hline
       \multirow{9}{*}{CLIP RN50}  &Lin. Probing & 85.67 &69.42&65.99\\ 
         &CoOp \cite{coop}&  89.88&74.04&69.04\\ 
        &CoCoOp \cite{cocoop}  & 88.69&74.80&69.48\\ 
        &CLIP-Adapter \cite{clip-adapter} & 88.86&75.31&69.29\\ 
         &DPL \cite{amortized} &89.24&74.86&69.10\\
         &Prograd \cite{prograd}& 88.51&75.40&69.49\\ 
         &TPT \cite{tpt}& 88.93 & 75.02 & 69.58 \\ 
         &\cellcolor{LightCyan}\textsc{StyLIP}-con& \cellcolor{LightCyan}90.56 & \cellcolor{LightCyan}75.92 & \cellcolor{LightCyan}70.41  \\ 
         &\cellcolor{LightCyan}\textsc{StyLIP}-sty& \cellcolor{LightCyan}91.77 & \cellcolor{LightCyan}76.39 & \cellcolor{LightCyan}70.87 \\ 
        &\cellcolor{LightCyan}\textsc{StyLIP}$^{*}$ & \cellcolor{LightCyan}89.14 & \cellcolor{LightCyan}75.67 & \cellcolor{LightCyan}69.85 \\  
        &\cellcolor{LightCyan}\textsc{StyLIP} &\cellcolor{LightCyan}\textbf{92.61} & \cellcolor{LightCyan}\textbf{77.18}&\cellcolor{LightCyan}\textbf{71.60}\\ 
         \midrule

           \multirow{9}{*}{CLIP ViT-B/16}  &Lin. Probing & 89.85 &76.15&77.71\\ 
         &CoOp \cite{coop}&  95.59&80.10&80.44\\ 
        &CoCoOp \cite{cocoop}  & 94.92&80.44&81.19\\ 
        &CLIP-Adapter \cite{clip-adapter} & 94.60&80.27&80.86\\ 
         &DPL \cite{amortized} &94.70&80.58&80.79\\
         &Prograd \cite{prograd}& 94.82&80.38&81.37\\ 
         &TPT \cite{tpt}&95.14 & 80.57 &  81.43\\ 
         ~ &MaPLe\cite{maple} & 95.33 & 80.15 & 81.95 \\
         &\cellcolor{LightCyan}\textsc{StyLIP}-con& \cellcolor{LightCyan}96.17 & \cellcolor{LightCyan}81.26 & \cellcolor{LightCyan}82.79  \\ 
         &\cellcolor{LightCyan}\textsc{StyLIP}-sty& \cellcolor{LightCyan}96.58 & \cellcolor{LightCyan}82.41 & \cellcolor{LightCyan}83.58 \\ 
        &\cellcolor{LightCyan}\textsc{StyLIP}* & \cellcolor{LightCyan}95.64 & \cellcolor{LightCyan}80.82 & \cellcolor{LightCyan}82.04 \\ 
        &\cellcolor{LightCyan}\textsc{StyLIP} &\cellcolor{LightCyan}\textbf{97.03} & \cellcolor{LightCyan} \textbf{82.90}&\cellcolor{LightCyan}\textbf{83.89}\\ 
        \bottomrule
    \end{tabular}}
    \vspace{-0.5cm}
    \label{tab:ssdg}
    }
\end{table}

\begin{table}[!ht]
\centering
\scriptsize{
    \centering
    \caption{Analysis of the generalization from base to new classes across domains. We show results on DomainNet with \textit{ClipArt} acting as the source domain, while others denote the target. The model is trained (backbone CLIP ViT-B/16) using $16$ shots from the base classes. (In $\%$)}
    \vspace{-0.32cm}
    \scalebox{0.68}{
    \begin{tabular}{lccccccc} 
    \toprule
     \multirow{4}{*}{\textbf{Method}}&\multicolumn{7}{c}{\textbf{DomainNet}}\\\cmidrule(lr){2-8}
    &\multicolumn{1}{c}{\textbf{Base}}&\multicolumn{6}{c}{\textbf{New}}\\\cmidrule(lr){2-2}\cmidrule(lr){3-8}
        ~ &\textit{\textbf{Clip Art}}&\textbf{Clip Art}&\textbf{Infograph}&\textbf{Painting}&\textbf{Quick Draw}&\textbf{Real}&\textbf{Sketch}\\ 
        \midrule 
        CLIP \cite{clip}&78.00&76.55&49.80&70.84&17.56&\textbf{88.11}&66.54\\ 
        CoOp \cite{coop}&82.79&75.60&48.60&71.38&20.90&85.19&67.39\\ 
        CoCoOp \cite{cocoop}&82.85&77.40&52.61&72.06&20.80&88.00&68.12\\ 
         CLIP-Adapter \cite{clip-adapter}&80.51&76.33&51.70&71.81&20.15&87.30&67.60\\ 
         DPL \cite{amortized} &82.35&76.49&52.10&71.88&20.30&87.54&67.73\\ 
         ProGrad \cite{prograd}& 83.00&77.50  &51.44  &72.16  & 20.86 &87.11&67.05\\ 
         MaPLe\cite{maple} & 82.84 & 77.61 & 51.65 & 71.95 & 20.51 & 87.35 & 67.82 \\
          \cellcolor{LightCyan}\textsc{StyLIP}-con& \cellcolor{LightCyan}83.71&\cellcolor{LightCyan}77.86 &\cellcolor{LightCyan}52.04  &\cellcolor{LightCyan}72.53  &\cellcolor{LightCyan} 20.97 &\cellcolor{LightCyan}87.20&\cellcolor{LightCyan}67.70\\ 
          
          \cellcolor{LightCyan}\textsc{StyLIP}-sty& \cellcolor{LightCyan}84.19&\cellcolor{LightCyan}77.62  &\cellcolor{LightCyan}52.80  &\cellcolor{LightCyan}73.00  &\cellcolor{LightCyan} 21.10 &\cellcolor{LightCyan}87.54&\cellcolor{LightCyan}68.29\\ 
          
        \cellcolor{LightCyan}\textsc{StyLIP}*& \cellcolor{LightCyan}83.34&\cellcolor{LightCyan}77.30  &\cellcolor{LightCyan}51.92  &\cellcolor{LightCyan}72.36  &\cellcolor{LightCyan} 20.93 &\cellcolor{LightCyan}87.39&\cellcolor{LightCyan}67.97\\ 
         
       \cellcolor{LightCyan}\textsc{StyLIP}&\cellcolor{LightCyan}\textbf{84.90}&\cellcolor{LightCyan}\textbf{78.14}&\cellcolor{LightCyan}\textbf{53.09}&\cellcolor{LightCyan}\textbf{73.60}&\cellcolor{LightCyan}\textbf{21.69}&\cellcolor{LightCyan}87.90&\cellcolor{LightCyan}\textbf{68.61}\\ \bottomrule
    \end{tabular}}\label{tab:b2n}}
    \vspace{-0.4cm}
\end{table}

\label{sec:Results}
\subsection{Comparison with state-of-the-art} 
We discuss the experimental comparisons of \textsc{StyLIP} with the literature in the following order of the DG tasks: \textbf{i)} multi-source DG, \textbf{ii)} single-source DG, \textbf{iii)} cross-domain base to novel class DG, \textbf{iv)} in-domain base to novel class DG and, \textbf{v)} cross-dataset DG, respectively. We follow the leave-one-domain-out evaluation protocol for multi-source DG where all the domains except one are considered source domains while the model is to be verified on the held-out target domain. For single-source DG, we train the model on one domain and test it on the remaining domains (leave-all-but-one-domain-out). We consider the standard few-shot training dataset with $16$-shots, following the CLIP literature \cite{coop, maple} for all the tasks. However, we have mentioned a detailed sensitivity analysis of \textsc{StyLIP} against the number of available training samples in Fig. \ref{fig_context}.

\noindent\textbf{Discussions on multi-source and single-source DG:} We present the mean leave-out performance of PACS, VLCS, Office-Home, Digits-DG, and DomainNet in Table \ref{tab_PACS} for both RN50 and ViT-B/16 backbones. Our method, \textsc{StyLIP}, surpasses zero-shot CLIP, Linear Probing, and domain alignment approaches by at least 3\% for both backbones, achieving state-of-the-art (SOTA) results. Additionally, \textsc{StyLIP} outperforms competitors, including DPL, CSVPT, and other prompting methods, across all datasets and vision backbones. Notably, when using ViT-B/16, \textsc{StyLIP} achieves outstanding performance with scores of 86.94\% on VLCS, 81.38\% on Digits-DG, and 62.03\% on DomainNet, surpassing others by at least 3\%.

Comparatively, the performance of \textsc{StyLIP}-con is slightly lower than \textsc{StyLIP} by approximately 0.5-1.3\%, while \textsc{StyLIP}-sty performs marginally better than \textsc{StyLIP}-con but remains inferior to \textsc{StyLIP}. However, both \textsc{StyLIP}-sty and \textsc{StyLIP}-con exhibit comparable or better performance than other prompting methods. The limitations of these variants of \textsc{StyLIP} are that they only capture partial visual properties, leading to sub-optimal prompt learning. In contrast, \textsc{StyLIP} fully leverages both style and content information of images, reducing the gap between visual and semantic spaces. Moreover, \textsc{StyLIP} outperforms \textsc{StyLIP}* due to the multi-scale content features, which are more generalizable than deeper semantically oriented visual representations.

In the single-source DG setting, using PACS, VLCS, and Office-Home datasets, we report the average leave-all-but-one-domain-out in Table \ref{tab:ssdg} for all domain combinations. Remarkably, \textsc{StyLIP} achieves a convincing improvement of approximately 1.4-2.5\% over other prompting techniques, establishing a new SOTA for single-source DG. For detailed domain-wise results in both single-source and multi-source DG setups, please refer to \textsc{supplementary}.

\begin{table}[!ht]
    \centering
    \caption{Comparison with SOTA methods on base-to-new generalization. \textsc{StyLIP} shows better generalization performance over existing methods on 11 different recognition datasets on $16$-shots and a context length of four. HM is the harmonic mean. (In $\%$)}
    \vspace{-0.2cm}
    \scalebox{0.63}{
    \begin{tabular}{lccc}
    \toprule
   \multicolumn{4}{c}{\textbf{Average over 11 datasets}}\\\midrule
     
 Methods&\multicolumn{1}{c}{Base}&\multicolumn{1}{c|}{New}&\multicolumn{1}{c}{HM}\\
  CLIP \cite{clip}&\multicolumn{1}{c}{ 69.34}&\multicolumn{1}{c|}{ 74.22}&\multicolumn{1}{c}{71.70 }\\
  
    CoOp \cite{coop}&\multicolumn{1}{c}{ 82.69}&\multicolumn{1}{c|}{ 63.22}&\multicolumn{1}{c}{ 71.66}\\
CoCoOp \cite{cocoop}&\multicolumn{1}{c}{80.47 }&\multicolumn{1}{c|}{71.69 }&\multicolumn{1}{c}{ 75.83}\\
LASP \cite{LASP}&\multicolumn{1}{c}{82.70 }&\multicolumn{1}{c|}{74.90 }&\multicolumn{1}{c}{78.61}\\
MaPLe \cite{maple}&\multicolumn{1}{c}{ 82.28}&\multicolumn{1}{c|}{75.14 }&\multicolumn{1}{c}{ 78.55}\\
\midrule



\cellcolor{LightCyan}\textsc{StyLIP}-con&\multicolumn{1}{c}{\cellcolor{LightCyan}{82.64}}&\multicolumn{1}{c|}
{\cellcolor{LightCyan}{75.39}}&\multicolumn{1}{c}{\cellcolor{LightCyan}{78.85}}\\

\cellcolor{LightCyan}\textsc{StyLIP}-sty&\multicolumn{1}{c}{\cellcolor{LightCyan}{82.93}}&\multicolumn{1}{c|}
{\cellcolor{LightCyan}{75.67}}&\multicolumn{1}{c}{\cellcolor{LightCyan}{79.13}}\\

\cellcolor{LightCyan}\textsc{StyLIP}$^{*}$&\multicolumn{1}{c}{\cellcolor{LightCyan}{82.30}}&\multicolumn{1}{c|}
{\cellcolor{LightCyan}{75.24}}&\multicolumn{1}{c}{\cellcolor{LightCyan}{78.61}}\\

\cellcolor{LightCyan}\textsc{StyLIP}&\multicolumn{1}{c}{\cellcolor{LightCyan}\textbf{83.22}}&\multicolumn{1}{c|}
{\cellcolor{LightCyan}{\textbf{75.94}}}&\multicolumn{1}{c}{\cellcolor{LightCyan}\textbf{79.41}}\\
        \bottomrule
    \end{tabular}
    
    \label{tab:b2n_image}
    }
\end{table} 

\begin{table*}[!ht]
    \centering
    \caption{Comparison of \textsc{StyLIP} with the prompt benchmark methods for generalization across datasets. We train the model on ImageNet using $16$-shots with CLIP ViT-B/16 and test on $10$ other datasets. (In $\%$)}
    \vspace{-0.4cm}
    \scalebox{0.6}{
    \begin{tabular}{llccccccccccc}
    \toprule
        \multicolumn{1}{c}{\textbf{Method}} & \multicolumn{1}{c}{\textbf{Source}} & \multicolumn{11}{c}{\textbf{Target}}  \\ \cmidrule(lr){2-2} \cmidrule(lr){3-13}
        
        & {ImgNet.} & {C101} & {Pets} & {Cars} & {Flowers} & {Food} & {Aircraft} & {Sun397} & {DTD} & {EuroSAT} & {UCF101} & {Average} \\ 
        \midrule
        CoOp \cite{coop} & 71.51 & 93.70 & 89.14 & 64.51 & 68.71 & 85.30 & 18.47 & 64.15 & 41.92 & 46.39 & 66.55 & 63.88 \\ 
        CoCoOp \cite{cocoop}& 71.02 & 94.43 & 90.14 & 65.32 & 71.88 & 86.06 & 22.94 & 67.36 & 45.73 & 45.37 & 68.21 & 65.74\\
        MaPLe \cite{maple}& 70.72&	93.53&	90.49&	65.57&	72.23&	86.20&	24.74&	67.01&	46.49&	48.06&	68.69&	66.30\\


        

\cellcolor{LightCyan}\textsc{StyLIP}-con&
\cellcolor{LightCyan}{71.44}&
\cellcolor{LightCyan}{94.96}&	
\cellcolor{LightCyan}{90.75}&
\cellcolor{LightCyan}{66.83}&	
\cellcolor{LightCyan}{72.14}&	
\cellcolor{LightCyan}{87.56}&	
\cellcolor{LightCyan}{24.88}&	
\cellcolor{LightCyan}{67.45}&	
\cellcolor{LightCyan}{46.63}&	
\cellcolor{LightCyan}{47.72}&	
\cellcolor{LightCyan}{68.85}&
\cellcolor{LightCyan}{66.47}\\

\cellcolor{LightCyan}\textsc{StyLIP}-sty&
\cellcolor{LightCyan}{72.05}&
\cellcolor{LightCyan}{95.13}&	
\cellcolor{LightCyan}{91.44}&
\cellcolor{LightCyan}{67.02}&	
\cellcolor{LightCyan}{72.29}&	
\cellcolor{LightCyan}{88.31}&	
\cellcolor{LightCyan}{25.17}&	
\cellcolor{LightCyan}{67.92}&	
\cellcolor{LightCyan}{47.64}&	
\cellcolor{LightCyan}{48.09}&	
\cellcolor{LightCyan}{69.12}&
\cellcolor{LightCyan}{67.25}\\

\cellcolor{LightCyan}\textsc{StyLIP}$^{*}$&
\cellcolor{LightCyan}{70.93}&
\cellcolor{LightCyan}{93.87}&	
\cellcolor{LightCyan}{90.53}&
\cellcolor{LightCyan}{65.75}&	
\cellcolor{LightCyan}{72.00}&	
\cellcolor{LightCyan}{86.85}&	
\cellcolor{LightCyan}{24.63}&	
\cellcolor{LightCyan}{67.30}&	
\cellcolor{LightCyan}{46.53}&	
\cellcolor{LightCyan}{47.92}&	
\cellcolor{LightCyan}{68.74}&
\cellcolor{LightCyan}{66.41}\\

\cellcolor{LightCyan}\textsc{StyLIP}&\cellcolor{LightCyan}\textbf{72.30}&\cellcolor{LightCyan}\textbf{95.45}&	\cellcolor{LightCyan}\textbf{91.60}&	\cellcolor{LightCyan}\textbf{67.09}&	\cellcolor{LightCyan}\textbf{72.36}&	\cellcolor{LightCyan}\textbf{88.60}&	\cellcolor{LightCyan}\textbf{25.21}&	\cellcolor{LightCyan}\textbf{68.11}&	\cellcolor{LightCyan}\textbf{47.86}&	\cellcolor{LightCyan}\textbf{48.22}&	\cellcolor{LightCyan}\textbf{69.30}	&\cellcolor{LightCyan}\textbf{67.38}\\
\bottomrule
        
    \end{tabular}}
    \label{fig:dgad}
    \vspace{-0.2cm}
\end{table*}

\noindent\textbf{Generalizing across novel domains and categories:}
In this experiment on DomainNet, we consider \texttt{ClipArt} as the source domain while the others denote the target domain. We divide the classes equally, and the model is trained and tested on the disjoint class sets, following \cite{cumix}.
In Table \ref{tab:b2n}, \textsc{StyLIP} outperforms the other prompting techniques in nine out of ten cases by $\approx 0.3-4 \%$ while generalizing to novel classes from both the source and the target domains, respectively. For the \textit{Real} domain of DomainNet, \textsc{StyLIP} lags \cite{cocoop} by a mere $0.21 \%$. \textsc{StyLIP} is less prone to overfitting to the classes of the source domain due to the better transferability offered by our model through multi-scale feature embedding. To validate this, we repeat this experiment using the model \textsc{StyLIP}*, which deals with only the deepest layer visual encodings.
Confirming our hypothesis, we find that the performance of \textsc{StyLIP}* is consistently poorer than \textsc{StyLIP} for all cases ($\approx 0.2-1.2 \%$).\\
\noindent\textbf{In-domain base to novel class generalization:} 
 In addition to the cross-domain generalization to novel categories, we show the performance of \textsc{StyLIP} on the $11$ datasets \cite{coop} where the base and novel classes are divided for each dataset to define the source and the target domains. A context length of four and $16$ samples per class are considered for training the model. 
\begin{figure}[t]
    \centering
    \includegraphics[width=0.49\columnwidth]{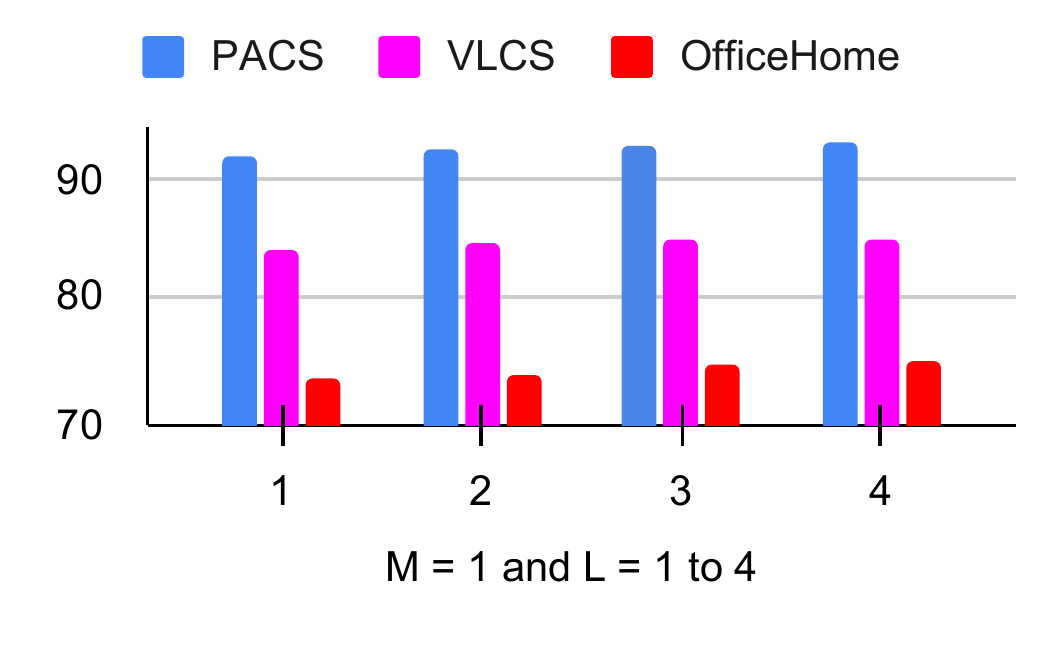}
    \includegraphics[width=0.49\columnwidth]{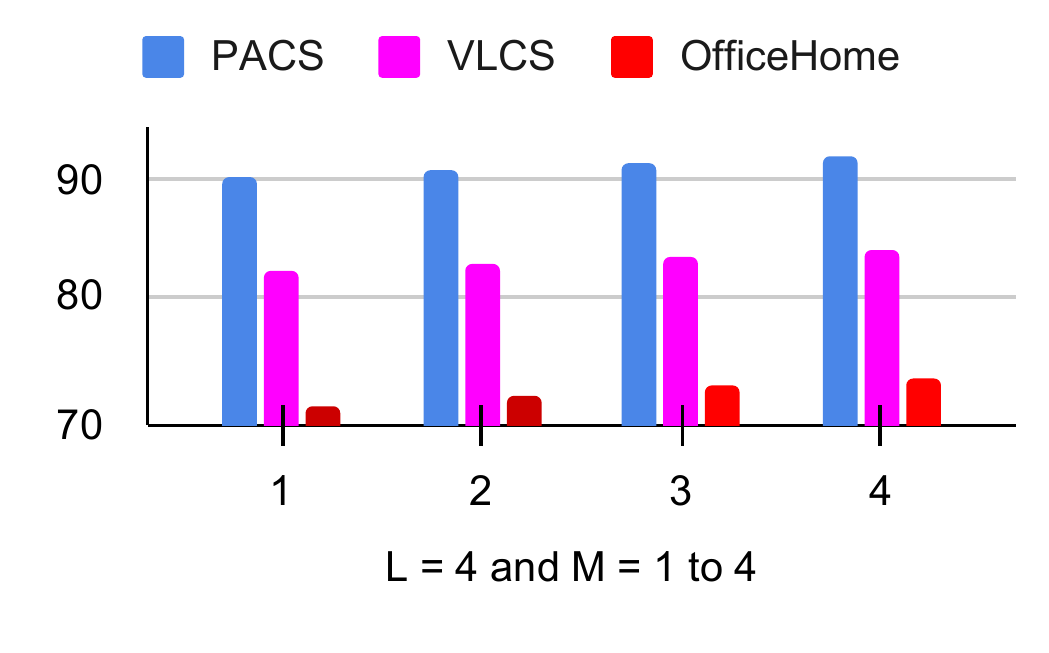}
    \vspace{-0.5cm}
    \caption{Effects of using style features from an increasing number of layers from $f_v$ for the fixed content features from the $L^{th}$ layer, and using content features from an increasing number of layers for a fixed style feature ($M=1$).}
    \vspace{-0.55cm}
    \label{fig:M_and_L}
\end{figure}
We depict the average performance over all the datasets in Table \ref{tab:b2n_image}, which shows that \textsc{StyLIP} beats the state-of-the-art, CoOp \cite{coop}, CoCoOp \cite{cocoop}, LASP \cite{LASP}, and MaPLe \cite{maple} convincingly by more than $0.8 \%$ on average H-score (H-score is the harmonic mean of the base-class and novel-class accuracies). Specifically, \textsc{StyLIP} is better than CoOp and CoCoOp by $\approx 8 \%$ and $4 \%$, respectively. We observe that \textsc{StyLIP} is able to beat the others both for the base as well as novel classes. This is important since the existing methods are mostly found to boost the performance of novel classes at the cost of decreasing base class performance. Refer to \texttt{Supplementary} for the detailed results.
\subsection{Ablation analysis}\label{sec:ablation}
\begin{figure}[ht!]
\vspace{-1.4cm}
\centering
   \includegraphics[width=\columnwidth]{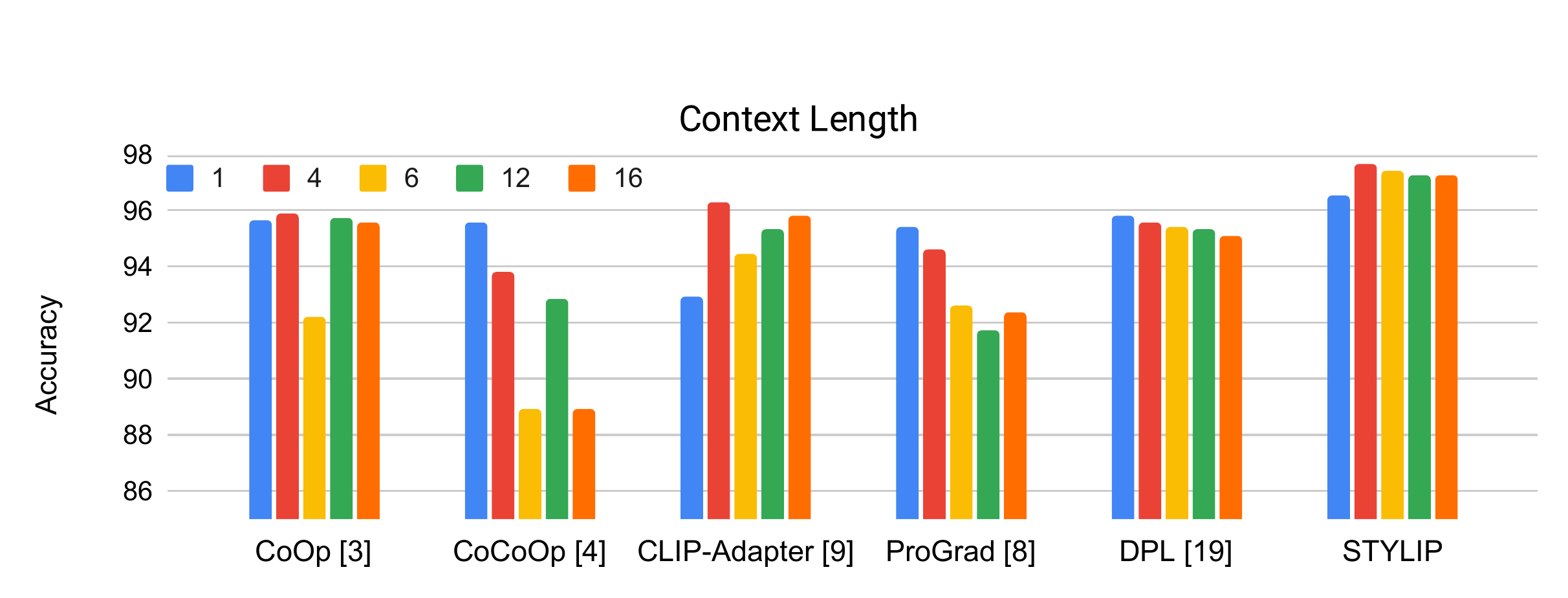}
   \includegraphics[width=\columnwidth]{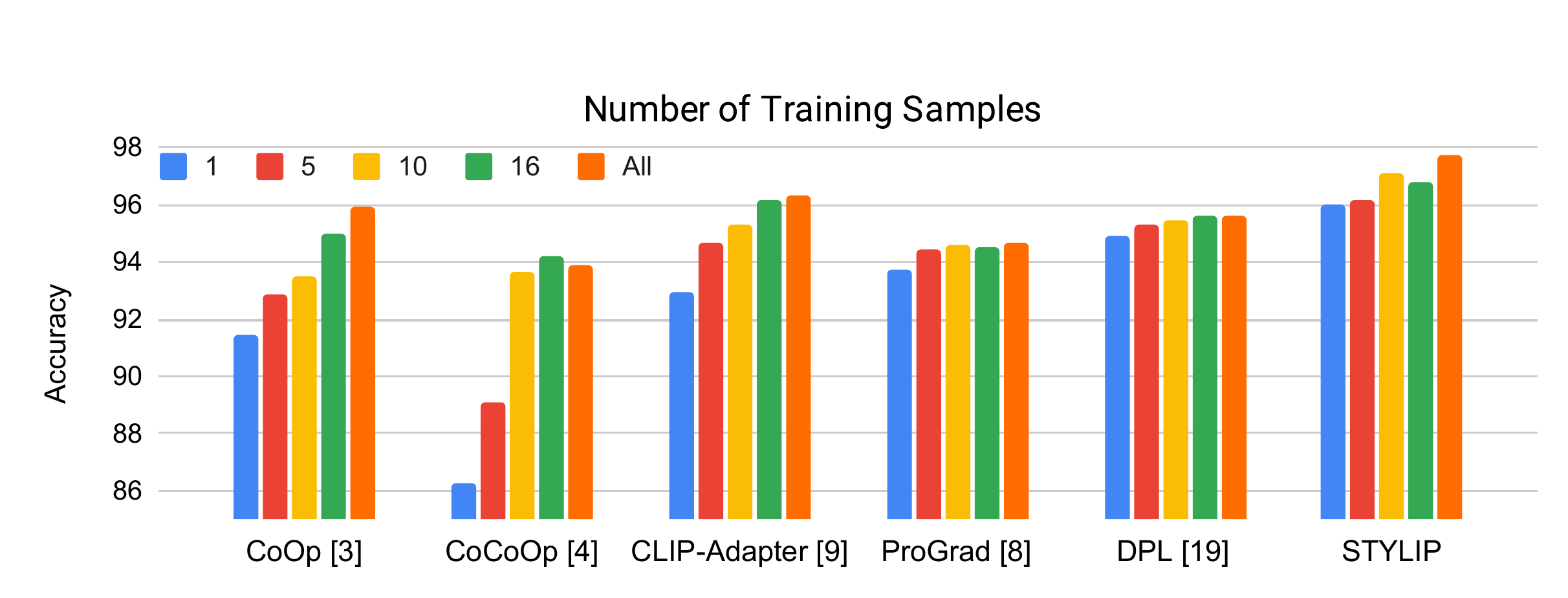}
\vspace{-0.9cm}
\caption{Sensitivity of prompting techniques (with ViT-B/16) on the context length ($M$) and the number of training samples per class. We show the average performance for multi-source DG on the PACS dataset.}
\label{fig_context}
\vspace{-0.3cm}
\end{figure}
\noindent\textbf{Generalization across datasets:}
Following the literature \cite{coop}, we perform prompt learning using $16$-shots from the $1000$ classes of ImageNet (source) and test on the other $10$ datasets (target). On the source domain, \textsc{StyLIP} beats the recent \cite{maple} by almost $1.6 \%$ (Table \ref{fig:dgad}). In contrast, for more specialized target datasets, such as DTD, EuroSAT and FGVAircrafts, \textsc{StyLIP} beats the other competitors. For the fine-grained datasets, \textsc{StyLIP} shows improvements up to $2 \%$, exhibiting much stronger transferability.\\
\noindent\textbf{Analysis of multi-scale features:} 
For the PACS and VLCS datasets with RN50, we conducted two experiments to investigate the impact of multi-scale features from $f_v$ on performance. In the first experiment, we focused on the content feature corresponding to the $L^{th}$ layer of $f_v$, varying $M$ from 1 to 4 for the style features.
In the second experiment, we fixed $M=1$ for the deepest layer style features and varied $L$ from 1 to 4 for the content features. As shown in Figure \ref{fig:M_and_L}, increasing the number of layers for both style and content features positively influenced the performance.

\begin{table}[!ht]
    \centering
    \small{
    \caption{Comparison of our proposed \textsc{StyLIP} with CoCoOp, including its extension with multi-scale features on the meta networks for PACS, VLCS, and Office-Home datasets for multi-source DG in terms of mean leave-one-out performance. (In $\%$)}
    \vspace{-0.4cm}
    \scalebox{0.65}{
    \begin{tabular}{clccc}
    \toprule
   \multirow{2}[1]{*}{\textbf{Backbone}}&\multirow{2}[1]{*}{\textbf{Method}}&\multicolumn{1}{c}{\textbf{PACS}}&\multicolumn{1}{c}{\textbf{VLCS}}&\multicolumn{1}{c}{\textbf{Office Home}}\\
          \cmidrule(lr){3-3}\cmidrule(lr){4-4}\cmidrule(lr){5-5}
       \midrule

       \multirow{5}{*}{CLIP RN50} 
        &CoCoOp \cite{cocoop}  & 88.69&74.80&69.48\\ 
        &MS-CoCoOp &89.15 &75.90 &69.72 \\
        &\cellcolor{LightCyan}\textsc{StyLIP} &\cellcolor{LightCyan}\textbf{92.61} & \cellcolor{LightCyan}\textbf{77.18}&\cellcolor{LightCyan}\textbf{71.60}\\ 
         \midrule

           \multirow{5}{*}{CLIP ViT-B/16}
        &CoCoOp \cite{cocoop}  & 94.92&80.44&81.19\\ 
        &MS-CoCoOp &95.14 &81.06 &81.93 \\
        &\cellcolor{LightCyan}\textsc{StyLIP} &\cellcolor{LightCyan}\textbf{97.03} & \cellcolor{LightCyan} \textbf{82.90}&\cellcolor{LightCyan}\textbf{83.89}\\ 
        \bottomrule
    \end{tabular}}
    \vspace{-0.3cm}
    \label{tab:ms}
    }
\end{table} 

In another experiment, we seek to show the usefulness of the multi-scale content features. In this regard, we compare \textsc{StyLIP} with a multi-scale version of CoCoOp \cite{cocoop} where we combine the multi-scale features to the input tokens instead of the deepest layer features as done in the CoCoOp paper. It can be observed from Tab. \ref{tab:ms} that \textsc{StyLIP} is able to beat MS-CoCoOp on multiple datasets for the Multi-DG task. This can be attributed to the improved prompting of \textsc{StyLIP} using the disentangled style and content features. 

    

\noindent\textbf{Context length $(M)$: }As we mention in Fig. \ref{fig_context}, we evaluate the effects of different context lengths for multi-source DG on PACS using the ViT backbone.
We find that \textsc{StyLIP} outperforms the other techniques, including \cite{cocoop,coop,prograd, amortized, clip-adapter} for context lengths of $1$, $4$, $6$, $12$, and $16$. To generate the style primitives for $M=16$, we choose to replicate the feature statistics vectors for the final four encoder layers, i.e., $\mathcal{F}_{9-12}(x)$, in addition to that of the original $12$ layers and feed them to $\{\mathcal{P}_m\}_{m=1}^{16}$.
A context of $4$ provides the optimal performance for \textsc{StyLIP}. We further find that a longer context length drastically deteriorates the performance of \cite{cocoop, prograd}, while \textsc{StyLIP} performs consistently across all context lengths. \\
\noindent\textbf{Sensitivity to the number of training samples:} To assess the robustness of \textsc{StyLIP} versus the number of training samples for the conventional DG setting, we train the single-source DG model on PACS while varying the number of training samples per class in the range $[1, 5, 10, 16, All]$. As shown in Fig. \ref{fig_context}, the DG performance of \cite{coop, cocoop} degrades in the low-data regime, while \cite{amortized, prograd, clip-adapter} shows comparatively better performance. Finally, \textsc{StyLIP} maintains its superior performance for very few training samples and shows improvements with more shots.

\begin{table}[t]
    \centering
    \caption{Ablation analysis of \textsc{StyLIP} for multi-source DG on PACS and Office-Home (OH) using ViT-B/16 backbone. (In $\%$)}
    \vspace{-0.33cm}
    \scalebox{0.62}{
    \begin{tabular}{l|cc}
    \toprule
        \multicolumn{1}{l}{\textbf{Baselines}}&\multicolumn{1}{c}{\textbf{PACS}}&\multicolumn{1}{c}{\textbf{OH}}\\
        \midrule
       \multicolumn{1}{l}{ Late Fusion Projector (max-pool)}&95.33&82.10\\
        \multicolumn{1}{l}{Late Fusion Projector (average pool)}&93.21&81.33\\
        \midrule
        \multicolumn{1}{l}{Depth of $\{\mathcal{P}_m\}_{m=1}^M$ and $\mathcal{P}_C$ (2 Layers)}&97.28&84.47\\
        \multicolumn{1}{l}{Depth $\{\mathcal{P}_m\}_{m=1}^M$ and $\mathcal{P}_C$ (3 Layers)}&97.43&84.40\\
        \midrule
        \multicolumn{1}{l}{$\{\mathcal{P}_m\}_{m=1}^M$ Only ($\mu$)}&96.81&83.26\\
        \multicolumn{1}{l}{$\{\mathcal{P}_m\}_{m=1}^M$ Only ($\sigma$)}&97.00&83.42\\
        \midrule
        \multicolumn{1}{l}{$\{\mathcal{B}_l\}_{l=1}^L$ (GAP over the spatial dimensions of the feature-maps)}&97.64&83.99\\
        \multicolumn{1}{l}{$\{\mathcal{B}_l\}_{l=1}^L$ (only flatten)}&97.30&83.57\\
        \multicolumn{1}{l}{$\{\mathcal{B}_l\}_{l=1}^L$ (conv 1x1 with $\hat{C}=2$)}&97.57&83.79\\
        \multicolumn{1}{l}{$\{\mathcal{B}_l\}_{l=1}^L$ (conv 1x1 with $\hat{C}=4$)}&97.92&84.33\\
        \multicolumn{1}{l}{$\{\mathcal{B}_l\}_{l=1}^L$ (conv 1x1 with $\hat{C}=16$)}&96.44&83.10\\
        \midrule
        \multicolumn{1}{l}{\cellcolor{LightCyan}\textsc{StyLIP} ($\hat{C}=3$)}&\cellcolor{LightCyan}\textbf{98.05}&\cellcolor{LightCyan}\textbf{84.63}\\
        \bottomrule
    \end{tabular}}
    \label{tab:ablation}
    \vspace{-0.55cm}
\end{table}

\noindent\textbf{Depth of style and content projectors:} To check the sensitivity of \textsc{StyLIP} on the depth of $\{\mathcal{P}_m\}_{m=1}^M$ and $\mathcal{P}_C$, we consider cases of multi-source DG where the projectors are two-layers and three-layers deep with a consistent number of nodes per layer, respectively, in PACS and Office-Home (Tab. \ref{tab:ablation}). We find that performance decreases marginally with increasing depth: $0.6-0.8 \%$ for PACS and $0.2-0.4 \%$ for Office-Home than \textsc{StyLIP} with linear projectors, suggesting \textsc{StyLIP} is indeed lightweight.\\
\noindent\textbf{Learnable vs. non-learnable $\mathcal{P}_A$}: Typically, $\mathcal{P}_C$ and $f_t$ produce feature embeddings of similar dimensions; hence, one way to fuse them in $\mathcal{P}_A$ is through element-wise feature pooling. In this regard, we use the max and average feature pooling strategies and observe in Tab. \ref{tab:ablation} that such aggregations affect the performance, reducing the multi-source DG accuracies on PACS and Office-Home by $\approx 2-3 \%$ in max pooling and $\approx 4-5 \%$ in average pooling than \textsc{StyLIP}. \\
\noindent\textbf{Analysis of style features}: Typically, the mean and std. of the feature maps together are known to capture the visual style information. To validate the same, we study the model's performance with either mean or std. being used as input to the style projectors. In this regard, we see a decrease in the performance of $1-2 \%$ compared to \textsc{StyLIP}, suggesting the importance of both statistical estimates. Interestingly, we see better accuracy when only std. is used for context learning than only mean (Tab. \ref{tab:ablation}).\\
\noindent\textbf{Sensitivity to the depth of the bottleneck layer $\hat{C}$}: We consider different $\hat{C}$ values in the range ${2, 3, 4, 16}$ to see the effects of the bottleneck dimensions in the final accuracy (Tab. \ref{tab:ablation}). While $\hat{C}=3$ provides the best performance, we see the numbers decreasing from $\hat{C}=4$ onwards, finally producing a dip of almost $1.5 \%$ for $\hat{C}=16$. Besides, we consider the scenario where $1 \times 1$ convolutions are not used, and we perform global average pooling (GAP), or directly flatten the feature maps and then concatenate. Both options perform poorly compared to \textsc{StyLIP} by $0.5-1 \%$.

\section{Takeaways}
\vspace{-0.2cm}
\label{sec:Conclusions}
In this paper, we aim to address the challenge of domain shift in DG tasks by proposing \textsc{StyLIP}, a domain-agnostic prompt learning strategy for CLIP. By disentangling and incorporating multi-scale visual style and content information from CLIP's frozen vision encoder into the prompt learning process, we enhance its generalizability. Extensive evaluations on various cross-domain inference tasks demonstrate the consistent state-of-the-art performance of \textsc{StyLIP}.
Our study on task-generalizable prompt learning paves the way for new research opportunities in computer vision. Future directions could explore domain-aware prompt learning with different foundation models and extend the proposed approach to structured prediction tasks.

{\small
\bibliographystyle{ieee_fullname}
\bibliography{egbib}
}

\end{document}


\title{Supplementary for \textsc{StyLIP}: Multi-Scale Style-Conditioned Prompt Learning\\for CLIP-based Domain Generalization}


\author{Shirsha Bose$^{1}\thanks{equal contribution}$ \and \hspace{-0.4cm}Ankit Jha$^{2*}$ \and \hspace{-0.4cm}Enrico Fini$^{3}$ \and \hspace{-0.3cm}Mainak Singha$^{2*}$ \and \hspace{-0.3cm}Elisa Ricci$^{3}$ \and \hspace{-0.4cm}Biplab Banerjee$^{2}$
\and
$^{1}$Technical University of Munich, Germany\and
$^{2}$Indian Institute of Technology Bombay, India\and
$^{3}$University of Trento, Italy
\and
{\tt\small shirshabosecs@gmail.com, ankitjha16@gmail.com, 
	enrico.fini@unitn.it}\and
 {\tt\small 
	mainaksingha.iitb@gmail.com, 
	e.ricci@unitn.it, getbiplab@gmail.com}}
\maketitle
\thispagestyle{empty}

We mention the following discussions in the supplementary:
\begin{itemize}
    \item Detailed dataset descriptions.
    \item Analysis of computational complexity.
    \item Detailed results for in-domain base-to-new class generalization over the $11$ datasets in Table \ref{tab:b2n_image}.
    \item Additional results on cross-domain base-to-new class generalization using \textit{ClipArt} as the source domain for the Office-Home dataset in Table \ref{tab:b2n}.
\end{itemize}

\subsection{Dataset details}
\label{sec:Experimenatl_protocol}
\ We evaluate \textsc{StyLIP} over five benchmark datasets for multi-source and single-source DG, which are described as follows: (1) \textbf{Office-Home} \cite{officehome} - It consists of 15,500 images coming from 65 classes covering four domains, namely, Art, Clipart, Product, and Real. (2) \textbf{PACS} \cite{dbadg} - Includes 9991 images consisting of seven classes that are spread across four domains, Artpaint, Cartoon, Sketch, and Photo.
(3) \textbf{VLCS} \cite{vlcs} - It was prepared by combining images from four image classification datasets, i.e., PASCAL VOC 2007 \cite{pascalvoc}, Caltech \cite{caltech}, LabelMe \cite{labelme}, and Sun \cite{sundataset}. It consists of images from five classes, Bird, Car, Chair, Dog, and Person. (4) \textbf{Digits-DG} \cite{l2a-ot} - This dataset is designed in the combination of handwritten digit recognition datasets, namely, MNIST \cite{mnist}, MNIST-M \cite{mnist-m}, SVHN \cite{svhn}, and SYN \cite{mnist-m}. (5) \textbf{DomainNet} \cite{domain_net} - It consists of images from six distinct domains, including real, painting, clipart, quickdraw, infograph, and sketch. Each domain has 48K - 172K images (600K in total) categorized into 345 classes. 

We further analyse the performance of \textsc{StyLIP} for cross dataset generalization, where \textsc{StyLIP} is trained on ImageNet \cite{krizhevsky2012imagenet} and tested on 10 other different datasets, including Caltech101 \cite{caltech}, OxfordPets \cite{oxford}, StanfordCars \cite{stanford}, Flowers102 \cite{flower}, Food101 \cite{food}, FGVCAircraft \cite{aircraft}, SUN397 \cite{sundataset}, DTD \cite{dtd}, EuroSAT \cite{helber2019eurosat} and UCF101 \cite{ucf}.

\noindent\textbf{Computation Complexity.}
We run our model on NVIDIA RTX 3090 Ti with 24 GB card. Tab. \ref{tab:compute} represents the comparison of computational complexity between different prompting techniques (CoOp \cite{coop}, CoCoOp \cite{cocoop}, and MaPLe \cite{maple}) in terms of GFLOPS relative to CoOp. MaPLe requires $0.12\%$ more computational overhead than CoOp and CoCoOp, whereas \textsc{StyLIP} needs $0.18 \%$ more resources than MaPLe, but \textsc{StyLIP} outperforms state-of-the-art MaPLe on the cross-dataset generalization (average over 11 datasets) approximately by $1.2\%$. 

    

\begin{table}[!ht]
    \centering
    \caption{Increase in compute w.r.t. CoOp and CoCoOp.}
    \scalebox{1}{
    \begin{tabular}{cccc}
    \toprule
     CoOp \cite{coop} & CoCoOp \cite{cocoop} & MaPLe \cite{maple} & \textsc{StyLIP} \\
    \midrule

    1$\times$ & 1$\times$ & +0.12$\%$& +0.18$\%$ \\
    
    \bottomrule
    \end{tabular}}
    \label{tab:compute}
\end{table}

\begin{table*}[!ht]
\scriptsize{
    \centering
    \caption{Comparison with state-of-the-art methods on base-to-new generalization. \textsc{StyLIP} shows better generalization performance over existing methods on 11 different recognition datasets on $16$-shots and a context length of four. HM represents the harmonic mean. (In $\%$)}
    \scalebox{1.2}{
    \begin{tabular}{lcccllcccllccc} 
     \multicolumn{4}{c}{(a) \textbf{Average over 11 datasets}}&\multicolumn{4}{c}{(b) ImageNet}&\multicolumn{4}{c}{(c) Caltech101}\\\cmidrule(lr){1-4}\cmidrule(lr){5-8}\cmidrule(lr){9-12}
     
 &\multicolumn{1}{c}{Base}&\multicolumn{1}{c|}{New}&\multicolumn{1}{c}{HM}&&\multicolumn{1}{c}{Base}&\multicolumn{1}{c|}{New}&\multicolumn{1}{c}{HM}&&\multicolumn{1}{c}{Base}&\multicolumn{1}{c|}{New}&\multicolumn{1}{c}{HM}\\\cmidrule(lr){1-4}\cmidrule(lr){5-8}\cmidrule(lr){9-12}

  CLIP \cite{clip}&\multicolumn{1}{c}{ 69.34}&\multicolumn{1}{c|}{ 74.22}&\multicolumn{1}{c}{71.70 }&CLIP \cite{clip}&\multicolumn{1}{c}{ 72.43}&\multicolumn{1}{c|}{68.14 }&\multicolumn{1}{c}{70.22 }&CLIP \cite{clip}&\multicolumn{1}{c}{96.84 }&\multicolumn{1}{c|}{ 94.00}&\multicolumn{1}{c}{ 95.40}\\
  
    CoOp \cite{coop}&\multicolumn{1}{c}{ 82.69}&\multicolumn{1}{c|}{ 63.22}&\multicolumn{1}{c}{ 71.66}&CoOp \cite{coop}&\multicolumn{1}{c}{76.47 }&\multicolumn{1}{c|}{67.88 }&\multicolumn{1}{c}{71.92 }&CoOp \cite{coop}&\multicolumn{1}{c}{ 98.00}&\multicolumn{1}{c|}{89.81 }&\multicolumn{1}{c}{ 93.73}\\
CoCoOp \cite{cocoop}&\multicolumn{1}{c}{80.47 }&\multicolumn{1}{c|}{71.69 }&\multicolumn{1}{c}{ 75.83}&CoCoOp \cite{cocoop}&\multicolumn{1}{c}{ 75.98}&\multicolumn{1}{c|}{70.43 }&\multicolumn{1}{c}{ 73.10}&CoCoOp \cite{cocoop}&\multicolumn{1}{c}{ 97.76}&\multicolumn{1}{c|}{93.81 }&\multicolumn{1}{c}{95.84 }\\
LASP \cite{LASP}&\multicolumn{1}{c}{ 82.70}&\multicolumn{1}{c|}{74.90}&\multicolumn{1}{c}{ 78.61}&LASP \cite{LASP}&\multicolumn{1}{c}{76.20}&\multicolumn{1}{c|}{70.95}&\multicolumn{1}{c}{ 73.48}&LASP \cite{LASP}&\multicolumn{1}{c}{ 98.10}&\multicolumn{1}{c|}{94.24}&\multicolumn{1}{c}{96.16}\\
MaPLe \cite{maple}&\multicolumn{1}{c}{ 82.28}&\multicolumn{1}{c|}{75.14 }&\multicolumn{1}{c}{ 78.55}&MaPLe \cite{maple}&\multicolumn{1}{c}{76.66 }&\multicolumn{1}{c|}{ 70.54}&\multicolumn{1}{c}{ 73.47}&MaPLe \cite{maple}&\multicolumn{1}{c}{ 97.74}&\multicolumn{1}{c|}{94.36 }&\multicolumn{1}{c}{96.02 }\\\cmidrule(lr){1-4}\cmidrule(lr){5-8}\cmidrule(lr){9-12}
\cellcolor{LightCyan}\textsc{StyLIP}-con&\multicolumn{1}{c}{\cellcolor{LightCyan}{82.64}} & \multicolumn{1}{c|}{\cellcolor{LightCyan}{75.39}} & \multicolumn{1}{c}{\cellcolor{LightCyan}{78.85} }&\cellcolor{LightCyan}\textsc{StyLIP}-con&\multicolumn{1}{c}{\cellcolor{LightCyan}{76.81}} & \multicolumn{1}{c|}{\cellcolor{LightCyan}{70.74}} & \multicolumn{1}{c}{\cellcolor{LightCyan}{73.65}} &\cellcolor{LightCyan}\textsc{StyLIP}-con& \multicolumn{1}{c}{\cellcolor{LightCyan}{97.74}} & \multicolumn{1}{c|}{\cellcolor{LightCyan}{94.83}} & \multicolumn{1}{c}{\cellcolor{LightCyan}{96.26}}   \\ 

\cellcolor{LightCyan}\textsc{StyLIP}-sty& \multicolumn{1}{c}{\cellcolor{LightCyan}{82.93}} & \multicolumn{1}{c|}{\cellcolor{LightCyan}{75.67}} & \multicolumn{1}{c}{\cellcolor{LightCyan}{79.13}} &\cellcolor{LightCyan}\textsc{StyLIP}-sty& \multicolumn{1}{c}{\cellcolor{LightCyan}{76.93}} & \multicolumn{1}{c|}{\cellcolor{LightCyan}{71.05}} & \multicolumn{1}{c}{\cellcolor{LightCyan}{73.87}}&\cellcolor{LightCyan}\textsc{StyLIP}-sty& \multicolumn{1}{c}{\cellcolor{LightCyan}{97.89}} & \multicolumn{1}{c|}{\cellcolor{LightCyan}{94.78}} & \multicolumn{1}{c}{\cellcolor{LightCyan}{96.31}} \\ 

\cellcolor{LightCyan}\textsc{StyLIP}^{*} & \multicolumn{1}{c}{\cellcolor{LightCyan}{82.30}} & \multicolumn{1}{c|}{\cellcolor{LightCyan}{75.24}} & \multicolumn{1}{c}{\cellcolor{LightCyan}{78.61}} &\cellcolor{LightCyan}\textsc{StyLIP}^{*} & \multicolumn{1}{c}{\cellcolor{LightCyan}{76.34}} & \multicolumn{1}{c|}{\cellcolor{LightCyan}{70.46}} & \multicolumn{1}{c}{\cellcolor{LightCyan}{73.28}} &\cellcolor{LightCyan}\textsc{StyLIP}^{*} & \multicolumn{1}{c}{\cellcolor{LightCyan}{97.45}} & \multicolumn{1}{c|}{\cellcolor{LightCyan}{94.61}} & \multicolumn{1}{c}{\cellcolor{LightCyan}{96.01}} \\ 

\cellcolor{LightCyan}\textsc{StyLIP}&\multicolumn{1}{c}{\cellcolor{LightCyan}\textbf{83.22}}&\multicolumn{1}{c|}
{\cellcolor{LightCyan}{\textbf{75.94}}}&\multicolumn{1}{c}{\cellcolor{LightCyan}\textbf{79.47}}& 
\cellcolor{LightCyan}\textsc{StyLIP}&\multicolumn{1}{c}{\cellcolor{LightCyan}\textbf{77.15}}&\multicolumn{1}{c|}{\cellcolor{LightCyan}\textbf{71.34}}&\multicolumn{1}{c}
{\cellcolor{LightCyan}\textbf{74.13}}&
\cellcolor{LightCyan}\textsc{StyLIP}&\multicolumn{1}{c}
{\cellcolor{LightCyan}\textbf{98.23} }&\multicolumn{1}{c|}{\cellcolor{LightCyan}\textbf{94.91}}&\multicolumn{1}{c}
{\cellcolor{LightCyan}\textbf{96.54}}\\
\cmidrule(lr){1-4}\cmidrule(lr){5-8}\cmidrule(lr){9-12}
&&&&&&&&&&&\\
 \multicolumn{4}{c}{(d) OxfordPets}&\multicolumn{4}{c}{(e) StanfordCars}&\multicolumn{4}{c}{(f) Flowers102}\\\cmidrule(lr){1-4}\cmidrule(lr){5-8}\cmidrule(lr){9-12}
     
 &\multicolumn{1}{c}{Base}&\multicolumn{1}{c|}{New}&
 \multicolumn{1}{c}{HM}&&\multicolumn{1}{c}{Base}&
 \multicolumn{1}{c|}{New}&\multicolumn{1}{c}{HM}&&
 \multicolumn{1}{c}{Base}&\multicolumn{1}{c|}{New}&
 \multicolumn{1}{c}{HM}\\\cmidrule(lr){1-4}\cmidrule(lr){5-8}\cmidrule(lr){9-12}

  CLIP \cite{clip}&\multicolumn{1}{c}{ 91.17}&
  \multicolumn{1}{c|}{97.26 }&\multicolumn{1}{c}{ 94.12}&
  CLIP \cite{clip}&\multicolumn{1}{c}{63.37 }&
  \multicolumn{1}{c|}{\textbf{74.89} }&\multicolumn{1}{c}{68.65 }&
  CLIP \cite{clip}&\multicolumn{1}{c}{72.08 }&\multicolumn{1}{c|}{\textbf{77.80}}&\multicolumn{1}{c}{74.83 }\\
  
    CoOp \cite{coop}&\multicolumn{1}{c}{93.67 }&
    \multicolumn{1}{c|}{ 95.29}&\multicolumn{1}{c}{ 94.47}&
    CoOp \cite{coop}&\multicolumn{1}{c}{ \textbf{78.12}}&
    \multicolumn{1}{c|}{60.40 }&\multicolumn{1}{c}{68.13}&
    CoOp \cite{coop}&\multicolumn{1}{c}{\textbf{97.60} }&
    \multicolumn{1}{c|}{ 59.67}&\multicolumn{1}{c}{74.06 }\\
    
CoCoOp \cite{cocoop}&\multicolumn{1}{c}{ 95.20}&
\multicolumn{1}{c|}{ 97.69}&\multicolumn{1}{c}{ 96.43}&
CoCoOp \cite{cocoop}&\multicolumn{1}{c}{70.49 }&
\multicolumn{1}{c|}{ 73.59}&\multicolumn{1}{c}{72.01 }&
CoCoOp \cite{cocoop}&\multicolumn{1}{c}{ 94.87}&
\multicolumn{1}{c|}{71.15 }&\multicolumn{1}{c}{81.71 }\\
LASP \cite{LASP}&\multicolumn{1}{c}{ 95.90}&\multicolumn{1}{c|}{97.93}&\multicolumn{1}{c}{ 96.90}&LASP \cite{LASP}&\multicolumn{1}{c}{75.17}&\multicolumn{1}{c|}{71.60}&\multicolumn{1}{c}{ 73.34}&LASP \cite{LASP}&\multicolumn{1}{c}{ 97.0}&\multicolumn{1}{c|}{74.0}&\multicolumn{1}{c}{83.95}\\

MaPLe \cite{maple}&\multicolumn{1}{c}{95.43 }&
\multicolumn{1}{c|}{97.76 }&\multicolumn{1}{c}{ 96.58}&
MaPLe \cite{maple}&\multicolumn{1}{c}{72.94 }&
\multicolumn{1}{c|}{74.00 }&\multicolumn{1}{c}{73.47 }&
MaPLe \cite{maple}&\multicolumn{1}{c}{95.92 }&
\multicolumn{1}{c|}{72.46 }&\multicolumn{1}{c}{ 82.56}\\
\cmidrule(lr){1-4}\cmidrule(lr){5-8}\cmidrule(lr){9-12}

\cellcolor{LightCyan}\textsc{StyLIP}-con&\multicolumn{1}{c}{\cellcolor{LightCyan}{95.66}} & \multicolumn{1}{c|}{\cellcolor{LightCyan}{97.94}} & \multicolumn{1}{c}{\cellcolor{LightCyan}{96.79}}&\cellcolor{LightCyan}\textsc{StyLIP}-con&\multicolumn{1}{c}{\cellcolor{LightCyan}{73.83}} & \multicolumn{1}{c|}{\cellcolor{LightCyan}{74.15}} & \multicolumn{1}{c}{\cellcolor{LightCyan}{73.99}} &\cellcolor{LightCyan}\textsc{StyLIP}-con& \multicolumn{1}{c}{\cellcolor{LightCyan}{96.14}} & \multicolumn{1}{c|}{\cellcolor{LightCyan}{72.75}} & \multicolumn{1}{c}{\cellcolor{LightCyan}{82.83}}   \\ 

\cellcolor{LightCyan}\textsc{StyLIP}-sty& \multicolumn{1}{c}{\cellcolor{LightCyan}{95.82}} & \multicolumn{1}{c|}{\cellcolor{LightCyan}{98.02}} & \multicolumn{1}{c}{\cellcolor{LightCyan}{96.91}} &\cellcolor{LightCyan}\textsc{StyLIP}-sty& \multicolumn{1}{c}{\cellcolor{LightCyan}{74.67}} & \multicolumn{1}{c|}{\cellcolor{LightCyan}{74.35}} & \multicolumn{1}{c}{\cellcolor{LightCyan}{74.51}}&\cellcolor{LightCyan}\textsc{StyLIP}-sty& \multicolumn{1}{c}{\cellcolor{LightCyan}{96.35}} & \multicolumn{1}{c|}{\cellcolor{LightCyan}{72.91}} & \multicolumn{1}{c}{\cellcolor{LightCyan}{83.01}} \\ 

\cellcolor{LightCyan}\textsc{StyLIP}^{*} & \multicolumn{1}{c}{\cellcolor{LightCyan}{95.57}} & \multicolumn{1}{c|}{\cellcolor{LightCyan}{97.82}} & \multicolumn{1}{c}{\cellcolor{LightCyan}{96.68}} &\cellcolor{LightCyan}\textsc{StyLIP}^{*} & \multicolumn{1}{c}{\cellcolor{LightCyan}{73.16}} & \multicolumn{1}{c|}{\cellcolor{LightCyan}{73.92}} & \multicolumn{1}{c}{\cellcolor{LightCyan}{73.54}} &\cellcolor{LightCyan}\textsc{StyLIP}^{*} & \multicolumn{1}{c}{\cellcolor{LightCyan}{96.02}} & \multicolumn{1}{c|}{\cellcolor{LightCyan}{72.53}} & \multicolumn{1}{c}{\cellcolor{LightCyan}{82.64}} \\ 

\cellcolor{LightCyan}\textsc{StyLIP}&\multicolumn{1}{c}{ \cellcolor{LightCyan}\textbf{95.96}}&\multicolumn{1}{c|}
{\cellcolor{LightCyan}\textbf{98.14} }&\multicolumn{1}{c}{ \cellcolor{LightCyan}\textbf{97.04}}&\cellcolor{LightCyan}
\textsc{StyLIP}&\multicolumn{1}{c}
{\cellcolor{LightCyan}75.19}&\multicolumn{1}{c|}
{ \cellcolor{LightCyan}74.46}&\multicolumn{1}{c}
{\cellcolor{LightCyan}\textbf{74.82} }&\cellcolor{LightCyan}
\textsc{StyLIP}&\multicolumn{1}{c}
{\cellcolor{LightCyan}96.54}&\multicolumn{1}{c|}
{\cellcolor{LightCyan}73.08}&\multicolumn{1}{c}
{\cellcolor{LightCyan}\textbf{83.19} }\\
\cmidrule(lr){1-4}\cmidrule(lr){5-8}\cmidrule(lr){9-12}
&&&&&&&&&&&\\
 \multicolumn{4}{c}{(g) Food101}&\multicolumn{4}{c}{(h) FGVCAircraft}&\multicolumn{4}{c}{(i) SUN397}\\\cmidrule(lr){1-4}\cmidrule(lr){5-8}\cmidrule(lr){9-12}
     
 &\multicolumn{1}{c}{Base}&\multicolumn{1}{c|}{New}
 &\multicolumn{1}{c}{HM}&&\multicolumn{1}{c}{Base}
 &\multicolumn{1}{c|}{New}&\multicolumn{1}{c}{HM}
 &&\multicolumn{1}{c}{Base}&\multicolumn{1}{c|}{New}
 &\multicolumn{1}{c}{HM}\\
 \cmidrule(lr){1-4}\cmidrule(lr){5-8}\cmidrule(lr){9-12}

  CLIP \cite{clip}&\multicolumn{1}{c}{90.10 }&
  \multicolumn{1}{c|}{ 91.22}&\multicolumn{1}{c}{90.66 }&
  CLIP \cite{clip}&\multicolumn{1}{c}{ 27.19}&
  \multicolumn{1}{c|}{\textbf{36.29}}&\multicolumn{1}{c}{31.09 }&
  CLIP \cite{clip}&\multicolumn{1}{c}{ 69.36}&
  \multicolumn{1}{c|}{ 75.35}&\multicolumn{1}{c}{ 72.23}\\
  
    CoOp \cite{coop}&\multicolumn{1}{c}{ 88.33}&
    \multicolumn{1}{c|}{ 82.26}&\multicolumn{1}{c}{ 85.19}&
    CoOp \cite{coop}&\multicolumn{1}{c}{ 40.44}&
    \multicolumn{1}{c|}{22.30 }&\multicolumn{1}{c}{28.75 }&
    CoOp \cite{coop}&\multicolumn{1}{c}{80.60 }&
    \multicolumn{1}{c|}{65.89 }&\multicolumn{1}{c}{ 72.51}\\
    
CoCoOp \cite{cocoop}&\multicolumn{1}{c}{90.70 }&
\multicolumn{1}{c|}{ 91.29}&\multicolumn{1}{c}{90.99}&
CoCoOp \cite{cocoop}&\multicolumn{1}{c}{ 33.41}&
\multicolumn{1}{c|}{ 23.71}&\multicolumn{1}{c}{27.74 }&
CoCoOp \cite{cocoop}&\multicolumn{1}{c}{79.74 }&
\multicolumn{1}{c|}{ 76.46}&\multicolumn{1}{c}{ 78.27}\\
LASP \cite{LASP}&\multicolumn{1}{c}{ 91.20}&\multicolumn{1}{c|}{91.70}&\multicolumn{1}{c}{ 91.44}&LASP \cite{LASP}&\multicolumn{1}{c}{34.53}&\multicolumn{1}{c|}{30.57}&\multicolumn{1}{c}{ 32.43}&LASP \cite{LASP}&\multicolumn{1}{c}{ 80.70}&\multicolumn{1}{c|}{78.60}&\multicolumn{1}{c}{79.63}\\

MaPLe \cite{maple}&\multicolumn{1}{c}{90.71 }&
\multicolumn{1}{c|}{ 92.05}&\multicolumn{1}{c}{91.38 }&
MaPLe \cite{maple}&\multicolumn{1}{c}{37.44 }&
\multicolumn{1}{c|}{35.61 }&\multicolumn{1}{c}{ 36.50}&
MaPLe \cite{maple}&\multicolumn{1}{c}{ 80.82}&
\multicolumn{1}{c|}{78.70 }&\multicolumn{1}{c}{ 79.75}\\
\cmidrule(lr){1-4}\cmidrule(lr){5-8}\cmidrule(lr){9-12}

\cellcolor{LightCyan}\textsc{StyLIP}-con&\multicolumn{1}{c}{\cellcolor{LightCyan}{90.92}} & \multicolumn{1}{c|}{\cellcolor{LightCyan}{92.23}} & \multicolumn{1}{c}{\cellcolor{LightCyan}{91.57} }&\cellcolor{LightCyan}\textsc{StyLIP}-con&\multicolumn{1}{c}{\cellcolor{LightCyan}{37.23}} & \multicolumn{1}{c|}{\cellcolor{LightCyan}{35.70}} & \multicolumn{1}{c}{\cellcolor{LightCyan}{36.45}} &\cellcolor{LightCyan}\textsc{StyLIP}-con& \multicolumn{1}{c}{\cellcolor{LightCyan}{81.23}} & \multicolumn{1}{c|}{\cellcolor{LightCyan}{78.94}} & \multicolumn{1}{c}{\cellcolor{LightCyan}{80.07}}   \\ 

\cellcolor{LightCyan}\textsc{StyLIP}-sty& \multicolumn{1}{c}{\cellcolor{LightCyan}{90.95}} & \multicolumn{1}{c|}{\cellcolor{LightCyan}{92.30}} & \multicolumn{1}{c}{\cellcolor{LightCyan}{91.62}} &\cellcolor{LightCyan}\textsc{StyLIP}-sty& \multicolumn{1}{c}{\cellcolor{LightCyan}{37.51}} & \multicolumn{1}{c|}{\cellcolor{LightCyan}{35.75}} & \multicolumn{1}{c}{\cellcolor{LightCyan}{36.61}}&\cellcolor{LightCyan}\textsc{StyLIP}-sty& \multicolumn{1}{c}{\cellcolor{LightCyan}{81.79}} & \multicolumn{1}{c|}{\cellcolor{LightCyan}{79.40}} & \multicolumn{1}{c}{\cellcolor{LightCyan}{80.58}} \\ 

\cellcolor{LightCyan}\textsc{StyLIP}^{*} & \multicolumn{1}{c}{\cellcolor{LightCyan}{90.84}} & \multicolumn{1}{c|}{\cellcolor{LightCyan}{92.11}} & \multicolumn{1}{c}{\cellcolor{LightCyan}{91.47}} &\cellcolor{LightCyan}\textsc{StyLIP}^{*} & \multicolumn{1}{c}{\cellcolor{LightCyan}{37.10}} & \multicolumn{1}{c|}{\cellcolor{LightCyan}{35.58}} & \multicolumn{1}{c}{\cellcolor{LightCyan}{36.32}} &\cellcolor{LightCyan}\textsc{StyLIP}^{*} & \multicolumn{1}{c}{\cellcolor{LightCyan}{80.95}} & \multicolumn{1}{c|}{\cellcolor{LightCyan}{78.80}} & \multicolumn{1}{c}{\cellcolor{LightCyan}{79.86}} \\ 

\cellcolor{LightCyan}\textsc{StyLIP}&
\multicolumn{1}{c}{\cellcolor{LightCyan}\textbf{91.20} }&
\multicolumn{1}{c|}{ \cellcolor{LightCyan}\textbf{92.48}}&
\multicolumn{1}{c}{\cellcolor{LightCyan}\textbf{91.84}}&
\cellcolor{LightCyan}\textsc{StyLIP}&
\multicolumn{1}{c}{\cellcolor{LightCyan}\textbf{37.65}}&
\multicolumn{1}{c|}{\cellcolor{LightCyan}35.93}&
\multicolumn{1}{c}{\cellcolor{LightCyan}\textbf{36.77}}&
\cellcolor{LightCyan}\textsc{StyLIP}&
\multicolumn{1}{c}{\cellcolor{LightCyan}\textbf{82.12}}&
\multicolumn{1}{c|}{\cellcolor{LightCyan}\textbf{79.95} }&
\multicolumn{1}{c}{\cellcolor{LightCyan}\textbf{81.02}}\\
\cmidrule(lr){1-4}\cmidrule(lr){5-8}\cmidrule(lr){9-12}
&&&&&&&&&&&&\\
 \multicolumn{4}{c}{(j) DTD}&\multicolumn{4}{c}{(k) EuroSAT}&\multicolumn{4}{c}{(l) UCF101}\\\cmidrule(lr){1-4}\cmidrule(lr){5-8}\cmidrule(lr){9-12}
     
 &\multicolumn{1}{c}{Base}&\multicolumn{1}{c|}{New}&
 \multicolumn{1}{c}{HM}&&\multicolumn{1}{c}{Base}&
 \multicolumn{1}{c|}{New}&\multicolumn{1}{c}{HM}&&
 \multicolumn{1}{c}{Base}&\multicolumn{1}{c|}{New}&
 \multicolumn{1}{c}{HM}\\
 \cmidrule(lr){1-4}\cmidrule(lr){5-8}\cmidrule(lr){9-12}

  CLIP \cite{clip}&\multicolumn{1}{c}{ 53.24}&
  \multicolumn{1}{c|}{ 59.90}&\multicolumn{1}{c}{56.37 }&
  CLIP \cite{clip}&\multicolumn{1}{c}{ 56.48}&
  \multicolumn{1}{c|}{ 64.05}&\multicolumn{1}{c}{60.03}&
  CLIP \cite{clip}&\multicolumn{1}{c}{70.53 }&
  \multicolumn{1}{c|}{77.50 }&\multicolumn{1}{c}{73.85 }\\
  
    CoOp \cite{coop}&\multicolumn{1}{c}{79.44 }&
    \multicolumn{1}{c|}{41.18 }&\multicolumn{1}{c}{54.24 }&
    CoOp \cite{coop}&\multicolumn{1}{c}{92.19 }&
    \multicolumn{1}{c|}{ 54.74}&\multicolumn{1}{c}{ 68.69}&
    CoOp \cite{coop}&\multicolumn{1}{c}{ 84.39}&
    \multicolumn{1}{c|}{ 56.05}&
    \multicolumn{1}{c}{67.46 }\\
    
CoCoOp \cite{cocoop}&\multicolumn{1}{c}{ 77.01}&
\multicolumn{1}{c|}{56.00 }&\multicolumn{1}{c}{64.85 }&
CoCoOp \cite{cocoop}&\multicolumn{1}{c}{87.49 }&
\multicolumn{1}{c|}{ 60.04}&\multicolumn{1}{c}{71.21 }&
CoCoOp \cite{cocoop}&\multicolumn{1}{c}{ 82.33}&
\multicolumn{1}{c|}{ 73.45}&\multicolumn{1}{c}{77.64 }\\
LASP \cite{LASP}&\multicolumn{1}{c}{ 81.40}&\multicolumn{1}{c|}{58.60}&\multicolumn{1}{c}{ 68.14}&LASP \cite{LASP}&\multicolumn{1}{c}{94.60}&\multicolumn{1}{c|}{77.78}&\multicolumn{1}{c}{ 85.36}&LASP \cite{LASP}&\multicolumn{1}{c}{ 84.77}&\multicolumn{1}{c|}{78.03}&\multicolumn{1}{c}{81.26}\\
MaPLe \cite{maple}&\multicolumn{1}{c}{ 80.36}&
\multicolumn{1}{c|}{ 59.18}&\multicolumn{1}{c}{68.16 }&
MaPLe \cite{maple}&\multicolumn{1}{c}{94.07 }&
\multicolumn{1}{c|}{73.23 }&\multicolumn{1}{c}{ 82.35}&
MaPLe \cite{maple}&\multicolumn{1}{c}{ 83.00}&
\multicolumn{1}{c|}{78.66 }&\multicolumn{1}{c}{80.77 }\\
\cmidrule(lr){1-4}\cmidrule(lr){5-8}\cmidrule(lr){9-12}

\cellcolor{LightCyan}\textsc{StyLIP}-con&\multicolumn{1}{c}{\cellcolor{LightCyan}{80.76}} & \multicolumn{1}{c|}{\cellcolor{LightCyan}{59.44}} & \multicolumn{1}{c}{\cellcolor{LightCyan}{68.48} }&\cellcolor{LightCyan}\textsc{StyLIP}-con&\multicolumn{1}{c}{\cellcolor{LightCyan}{94.45}} & \multicolumn{1}{c|}{\cellcolor{LightCyan}{73.67}} & \multicolumn{1}{c}{\cellcolor{LightCyan}{82.78}} &\cellcolor{LightCyan}\textsc{StyLIP}-con& \multicolumn{1}{c}{\cellcolor{LightCyan}{84.24}} & \multicolumn{1}{c|}{\cellcolor{LightCyan}{78.93}} & \multicolumn{1}{c}{\cellcolor{LightCyan}{81.50}}   \\ 

\cellcolor{LightCyan}\textsc{StyLIP}-sty& \multicolumn{1}{c}{\cellcolor{LightCyan}{81.23}} & \multicolumn{1}{c|}{\cellcolor{LightCyan}{60.94}} & \multicolumn{1}{c}{\cellcolor{LightCyan}{69.64}} &\cellcolor{LightCyan}\textsc{StyLIP}-sty& \multicolumn{1}{c}{\cellcolor{LightCyan}{94.57}} & \multicolumn{1}{c|}{\cellcolor{LightCyan}{73.85}} & \multicolumn{1}{c}{\cellcolor{LightCyan}{82.94}}&\cellcolor{LightCyan}\textsc{StyLIP}-sty& \multicolumn{1}{c}{\cellcolor{LightCyan}{84.51}} & \multicolumn{1}{c|}{\cellcolor{LightCyan}{79.05}} & \multicolumn{1}{c}{\cellcolor{LightCyan}{81.69}} \\ 

\cellcolor{LightCyan}\textsc{StyLIP}^{*} & \multicolumn{1}{c}{\cellcolor{LightCyan}{80.22}} & \multicolumn{1}{c|}{\cellcolor{LightCyan}{59.60}} & \multicolumn{1}{c}{\cellcolor{LightCyan}{68.39}} &\cellcolor{LightCyan}\textsc{StyLIP}^{*} & \multicolumn{1}{c}{\cellcolor{LightCyan}{94.33}} & \multicolumn{1}{c|}{\cellcolor{LightCyan}{73.46}} & \multicolumn{1}{c}{\cellcolor{LightCyan}{82.60}} &\cellcolor{LightCyan}\textsc{StyLIP}^{*} & \multicolumn{1}{c}{\cellcolor{LightCyan}{83.37}} & \multicolumn{1}{c|}{\cellcolor{LightCyan}{78.72}} & \multicolumn{1}{c}{\cellcolor{LightCyan}{80.98}} \\ 

\cellcolor{LightCyan}\textsc{StyLIP}&\multicolumn{1}{c}{ \cellcolor{LightCyan}\textbf{81.57}}&\multicolumn{1}{c|}{\cellcolor{LightCyan}\textbf{61.72}}&\multicolumn{1}{c}{ \cellcolor{LightCyan}\textbf{70.27}}&\cellcolor{LightCyan}\textsc{StyLIP}&\multicolumn{1}{c}{\cellcolor{LightCyan}\textbf{94.61}}&\multicolumn{1}{c|}{\cellcolor{LightCyan}\textbf{74.06}}&\multicolumn{1}{c}{\cellcolor{LightCyan}\textbf{83.08}}&\cellcolor{LightCyan}\textsc{StyLIP}&\multicolumn{1}{c}{\cellcolor{LightCyan}\textbf{85.19}}&\multicolumn{1}{c|}{\cellcolor{LightCyan}\textbf{79.22}}&\multicolumn{1}{c}{\cellcolor{LightCyan}\textbf{82.10}}\\

\cmidrule(lr){1-4}\cmidrule(lr){5-8}\cmidrule(lr){9-12}
    \end{tabular}}
    \label{tab:b2n_image}}
\end{table*}

\begin{table}[!ht]
\centering
\scriptsize{
    \centering
    \caption{Analysis of the generalization from base to new classes across domains. We show results on Office-Home with \textit{ClipArt} acting as the source domain, while others denote the target. The model is trained (backbone CLIP ViT-B/16) using $16$ shots from the base classes. (In $\%$)}
    \scalebox{0.8864}{
    \begin{tabular}{lccccc} 
    \toprule
     \multirow{4}{*}{\textbf{Method}}&\multicolumn{5}{c}{\textbf{Office-Home}}\\\cmidrule(lr){2-6}
    &\multicolumn{1}{c}{\textbf{Base}}&\multicolumn{4}{c}{\textbf{New}}\\\cmidrule(lr){2-2}\cmidrule(lr){3-6}
        ~ & \textit{\textbf{Clip Art}} & \textbf{Art} & \textbf{Clip Art} & \textbf{Product} & \textbf{Real World} \\ 
        \midrule 
        CLIP \cite{clip}  & 78.12 &62.01 & 77.78 & 87.52 & 88.02 \\ 
        CoOp \cite{coop}& 82.60 &70.60 & 82.23 & 90.44 & 87.21 \\ 
        CoCoOp \cite{cocoop}& 82.64 &71.00 &83.61 & 92.12 & 89.19\\ 
         CLIP-Adapter \cite{clip-adapter}& 80.00 &73.19 & 83.00 & 92.11 & 89.53 \\ 
         DPL \cite{amortized}&  82.20 &71.54 & 82.80 &92.37 & 89.15 \\ 
         ProGrad \cite{prograd}& 82.41 &72.00 &83.29 &92.11&89.585\\ 
         \cellcolor{LightCyan}\textsc{StyLIP}-con& \cellcolor{LightCyan}82.67&\cellcolor{LightCyan}72.10 &\cellcolor{LightCyan}84.39 &\cellcolor{LightCyan}92.17&\cellcolor{LightCyan}90.24\\ 
         \cellcolor{LightCyan}\textsc{StyLIP}-sty& \cellcolor{LightCyan}83.22&\cellcolor{LightCyan}72.60 &\cellcolor{LightCyan}84.51 &\cellcolor{LightCyan}92.78&\cellcolor{LightCyan}91.25\\ 
         
        \cellcolor{LightCyan}\textsc{StyLIP}*& \cellcolor{LightCyan}83.90&\cellcolor{LightCyan}73.48 &\cellcolor{LightCyan}85.07 &\cellcolor{LightCyan}92.60&\cellcolor{LightCyan}90.77\\ 
         
       \cellcolor{LightCyan}\textsc{StyLIP}&\cellcolor{LightCyan}\textbf{84.33} & \cellcolor{LightCyan}\textbf{74.60} & \cellcolor{LightCyan}\textbf{87.25}  & \cellcolor{LightCyan}\textbf{93.00} & \cellcolor{LightCyan}\textbf{91.42} \\ \bottomrule
    \end{tabular}}\label{tab:b2n}}
    
\end{table}

{\small
\bibliographystyle{ieee_fullname}
\bibliography{egbib}
}